\pdfoutput=1
\documentclass{article} %
\usepackage[preprint]{colm2026_conference}
\usepackage{microtype}
\usepackage{url}
\usepackage{caption}
\usepackage{subcaption}
\usepackage{multirow}
\usepackage{makecell}
\usepackage{colortbl}
\usepackage{array}
\usepackage{setspace}
\usepackage{booktabs}
\usepackage{graphicx}
\usepackage{ragged2e}
\usepackage{tabularx}
\usepackage{xltabular}
\usepackage{csvsimple}
\usepackage{longtable}
\usepackage{pifont}
\usepackage{enumitem}
\usepackage{amsmath}
\usepackage{hyperref}
\usepackage{cleveref}  %
\usepackage{wrapfig}
\usepackage[T1]{fontenc} 
\usepackage[utf8]{inputenc}
\usepackage[most]{tcolorbox}
\usepackage{threeparttable}
\usepackage{xurl}  
\tcbuselibrary{breakable, skins}
\usepackage{tablefootnote}

\usepackage{threeparttable}
\usepackage{booktabs}
\usepackage{makecell}
\usepackage{multirow}

\newcommand{\cmidruleFull}{%
    \noalign{\global\arrayrulewidth=0.9pt}%
    \cline{2-4}%
    \noalign{\global\arrayrulewidth=0.4pt}%
}

\newcommand{\cmark}{\ding{51}}  %
\newcommand{\xmark}{\ding{55}}  %

\makeatletter
\newcommand{\customcite}[2]{%
  \begingroup
  \NAT@swafalse
  \hyper@natlinkstart{#2}#1\hyper@natlinkend
  \endgroup
}
\makeatother

\usepackage{lineno}
\usepackage{indentfirst}

\definecolor{darkblue}{rgb}{0, 0, 0.5}
\definecolor{darkgray}{RGB}{64, 64, 62} %
\definecolor{rust}{RGB}{204, 120, 92}  
\definecolor{cloudlight}{RGB}{214, 214, 214}
\definecolor{yellow}{RGB}{235, 219, 188}
\definecolor{light_orange}{RGB}{212, 162, 127}
\definecolor{ivory}{RGB}{240,240,235}

\newcommand{\highlight}[1]{\colorbox{yellow}{#1}}

\newtcolorbox{PromptBox}[2][]{
    colback=white,
    colframe=rust,
    enhanced jigsaw,
    boxrule=2pt,
    arc=4mm,
    width=\textwidth,
    sharp corners=all,
    attach boxed title to top center={yshift=-2mm},
    boxed title style={colback=darkgray, colframe=rust, fontupper=\bfseries\color{white}},
    title={#2},
    before upper={\scriptsize\ttfamily\selectfont},
    #1,
}

\newtcolorbox{PromptBoxBreak}[2][]{
    colback=white,
    colframe=rust,
    enhanced,
    breakable,
    boxrule=2pt,
    arc=4mm,
    width=\textwidth,
    enlarge left by=0mm,
    enlarge right by=0mm,
    left=2pt,
    right=2pt,
    top=2pt,
    bottom=2pt,
    sharp corners=all,
    attach boxed title to top center={yshift=-2mm},
    boxed title style={colback=darkgray, colframe=rust, fontupper=\bfseries\color{white}},
    title={#2},
    before upper={\scriptsize\ttfamily\selectfont},
    #1,
}

\hypersetup{colorlinks=true, citecolor=darkblue, linkcolor=darkblue, urlcolor=darkblue}

\title{FrontierFinance: A Long-Horizon Computer-Use Benchmark of Real-World Financial Tasks}

\author{%
Michael Krumdick$^1$\thanks{Corresponding author: \texttt{michael.krumdick@kensho.com}} \;
Varshini Reddy$^1$ \;
Shivani Chaudhary$^1$ \;
William Day$^1$ \; \\
\textbf{Maarij Ahmed$^2$ \;
Hayan Haqqi$^2$ \;
Muhammad Ahsen Fahim$^2$\;} \\
\textbf{Hanzallah Amjad$^2$ \; 
Ahmad Orakzai$^2$ \;
Aqsa Gul$^2$
Chris Tanner$^{1,3}$} \\
$^1$Kensho Technologies, $^2$S\&P Global, $^3$MIT
}

\begin{document}

\ifcolmsubmission
\linenumbers
\fi

\maketitle

\begin{abstract}

As concerns surrounding AI-driven labor displacement intensify in knowledge-intensive sectors, existing benchmarks fail to measure performance on tasks that define practical professional expertise. Finance, in particular, has been identified as a domain with high AI exposure risk, yet lacks robust benchmarks to track real-world developments. This gap is compounded by the absence of clear accountability mechanisms in current Large Language Model (LLM) deployments. To address this, we introduce FrontierFinance, a long-horizon benchmark of 25 complex financial modeling tasks across five core finance models, requiring an average of over 18 hours of skilled human labor per task to complete. Developed with financial professionals, the benchmark reflects industry-standard financial modeling workflows and is paired with detailed rubrics for structured evaluation. We engage human experts to define the tasks, create rubrics, grade LLMs, and perform the tasks themselves as human baselines. We demonstrate that our human experts both receive higher scores on average, and are more likely to provide client-ready outputs than current state-of-the-art systems.

\end{abstract}

\section{Introduction}
The rapid integration of Large Language Models (LLMs) into the global economy has the potential to precipitate a significant shift in the labor market \citep{davenport2026companies}. Recent economic analyses indicate that the risk of displacement is most acute in cognitive-heavy professions---such as financial analysis, legal reasoning, and software engineering---where workflows are increasingly being restructured based on the potential of AI agents rather than on their verified performance \citep{goldman2026how, anthropic2026labor}. Despite these systemic shifts, there is a critical lack of empirical evidence regarding whether LLMs can truly replicate the high-fidelity, long-horizon reasoning required in these roles \citep{gimbel2025evaluating}. 

This evaluation gap stems from a fundamental limitation in current benchmarking paradigms. While LLMs have demonstrated exemplary performance on short-horizon reasoning tasks \citep{hendrycks2021measuringmassivemultitasklanguage, srivastava2023imitationgamequantifyingextrapolating}, most existing datasets emphasize narrow capabilities such as isolated question answering or simplified code generation \citep{zhao-etal-2024-knowledgefmath, krumdick-etal-2024-bizbench, reddy-etal-2024-docfinqa}. These benchmarks fail to capture the complexity of professional workflows that require sustained decision-making over extended sequences of sub-tasks and the synthesis of interdependent, time period varying documents \citep{liu2025agentbenchevaluatingllmsagents, jimenez2024swebenchlanguagemodelsresolve}. Consequently, current metrics may provide a false sense of security regarding model reliability in professional settings where errors propagate across complex systems \citep{measuring_ai_ability_metr}.

Financial modeling serves as a critical proxy for this broader labor risk. It is a quintessentially human-expert task, often requiring 5--20 hours of focused labor to construct a single artifact. Professional modeling---encompassing three-statement projections, leveraged buyouts (LBOs), lender model, Discounted Cash Flow (DCF) model and merger analyses---demands not only domain expertise but also the ability to maintain strict logical and mathematical consistency across thousands of interconnected spreadsheet cells. In this domain, a single error in a cell reference or formula dependency can cascade through thousands of interconnected cells and compromise the model's structural integrity. Despite the high-stakes nature of financial labor, the domain remains significantly underexplored in agentic benchmarking \citep{zhang-etal-2025-xfinbench}. These models directly inform high-stakes capital allocation decisions, where a single misspecified assumption, such as a one percentage point error in the discount rate, can shift implied valuation by hundreds of millions of dollars. Although the final decision incorporates qualitative factors such as legal diligence and market conditions, the financial model serves as the central quantitative artifact around which pricing, deal terms, and investment approvals are structured.

To address these gaps, we introduce \textbf{FrontierFinance}\footnote{Open source versions of the code and underlying data are forthcoming.}, a benchmark designed to evaluate LLMs on end-to-end financial modeling tasks across 5 core disciplines: three-statement, Leveraged Buyout (LBO), Discounted Cash Flow (DCF), merger, and lender models. Unlike existing financial datasets that focus on information retrieval, FrontierFinance requires the construction of complete, reconciled models from scratch based on detailed prompts.

\begin{figure*}[t]
    \centering
    \includegraphics[width=\textwidth]{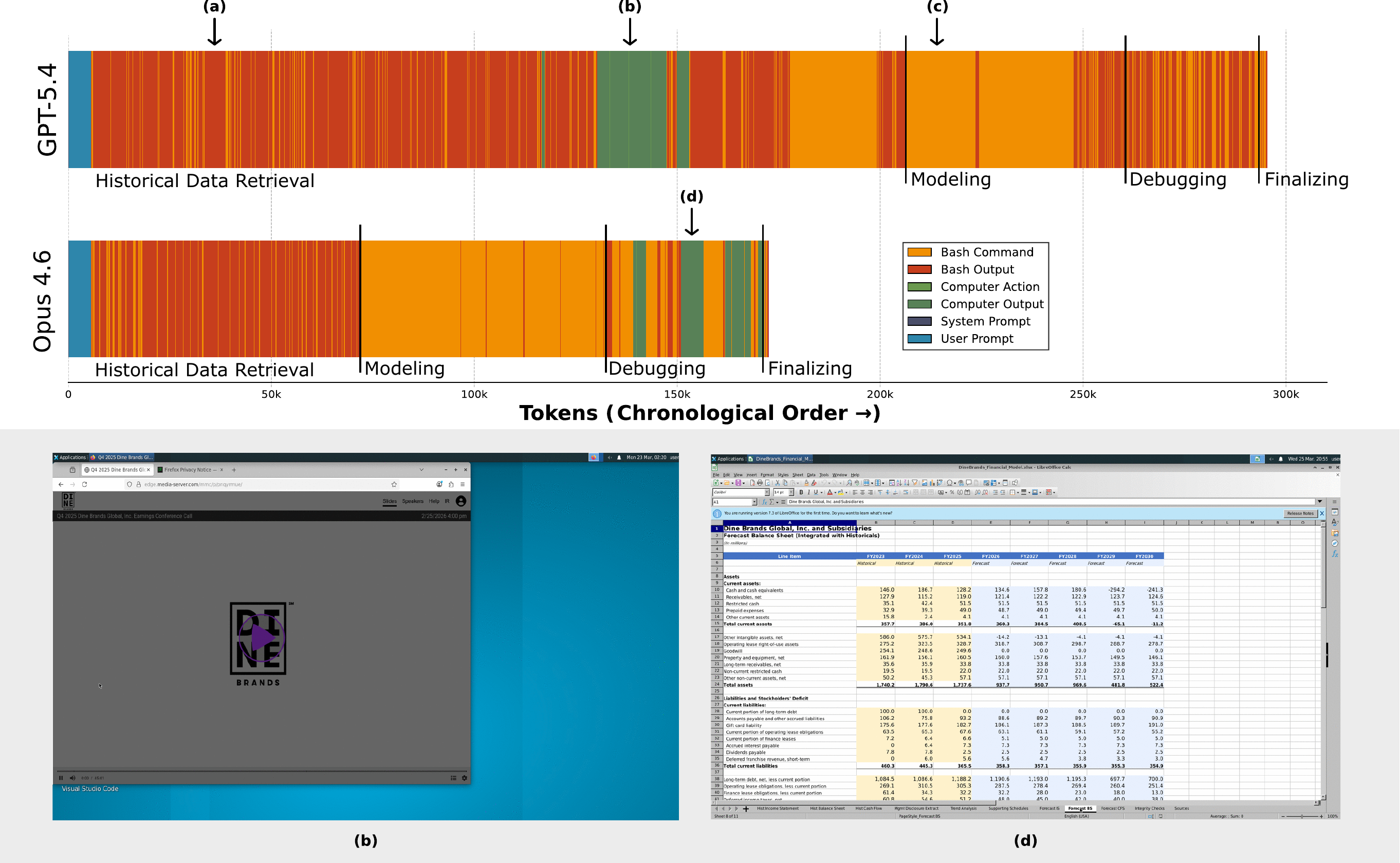}
    \caption{Comparison of two LLM approaches on the same Three-Statement Model. The bars represent the chronological token usage during execution. \textbf{(a)} corresponds to the data retrieval stage, where models issue short bash commands and receive longer outputs. \textbf{(b)} illustrates an example of GPT-5.4’s computer use, where it unsuccessfully attempts to access an earnings call video for additional context. \textbf{(c)} denotes the primary modeling phase, during which both models construct an initial version of the financial model, typically within a single large Python code block spanning hundreds of lines. \textbf{(d)} highlights a characteristic behavior of Opus 4.6, which frequently inspects generated spreadsheets using LibreOffice.}
    \label{fig:fig_test}
\end{figure*}

In addition to task design, we define a comprehensive rubric-based evaluation framework that enables systematic comparison between human and model-generated outputs. We collect human baselines to characterize the time and effort required to complete these tasks and compare them against LLM-based agents in terms of both execution time and output quality. To support consistent evaluation, we develop detailed, rubric-based scoring schemes tailored to each model type. We also incorporate an LLM-as-a-judge framework to scale evaluation while maintaining alignment with expert human judgment. We validate this approach by measuring agreement between automated scores and human annotations, demonstrating that carefully designed rubrics can enable reliable and efficient assessment of complex, structured outputs. 

\begingroup
\setlength{\tabcolsep}{4pt}
\renewcommand{\arraystretch}{1.1}
\begin{table*}[t]
\centering
\small
\begin{threeparttable}
\resizebox{\textwidth}{!}{%
\begin{tabular}{l l l c c c c c c}
\toprule
\multirow{2}{*}{\textbf{Benchmark}} &
\multirow{2}{*}{\makecell{\textbf{Total}\\ \textbf{Tasks}}} &
\multirow{2}{*}{\makecell{\textbf{\# Finance}\\ \textbf{Tasks}}} &
\multicolumn{2}{c}{\textbf{Avg. Human Hours (h)}} &
\multirow{2}{*}{\makecell{\textbf{Reference}\\ \textbf{Answer}}} &
\multirow{2}{*}{\makecell{\textbf{\# Unique}\\ \textbf{Output Formats}}} &
\multirow{2}{*}{\makecell{\textbf{Detailed}\\ \textbf{Rubric}}} &
\multirow{2}{*}{\makecell{\textbf{Source}\\ \textbf{Docs}\textsuperscript{A}}} \\
\cline{4-5} \addlinespace[2pt]
& & & \textbf{Estimated} & \textbf{Actual} & & & & \\
\midrule
\customcite{APEX--Agents}{vidgen2026apexagents} & 480 & 160 & 1.8 & 1.4 & \cmark & >2 & $\sim$ & \xmark \\
\customcite{GDPval}{patwardhan2025gdpvalevaluatingaimodel} & 220\textsuperscript{B} & 150 & 7 & - & \xmark & >2 & $\sim$ & \xmark \\
\customcite{TOOLATHLON}{li2026tooldecathlonbenchmarkinglanguage} & 108 & 10 & 5 & - & \xmark & 1 & \xmark & \cmark \\
\customcite{LongCLI-Bench}{feng2026longclibenchpreliminarybenchmarkstudy} & 20 & - & - & >17 & \cmark & 1 & \xmark & \xmark \\
\customcite{RE-Bench}{wijk2025rebenchevaluatingfrontierai} & 7 & - & - & 8\textsuperscript{C} & \cmark & 1 & \xmark & \xmark \\
\midrule
FrontierFinance & 25 & 25 & 12.5 & 18.3 & \cmark & 2 & \cmark & \cmark \\
\bottomrule
\end{tabular}
}
\caption{Comparison of existing long-horizon LLM benchmarks. ``Reference answer'' indicates whether reference solutions, if any, are publicly available.\\
\footnotesize \textsuperscript{A} Agents are expected to independently source all relevant files to complete the tasks from the internet.\\
\textsuperscript{B} Benchmark reports 1,320 total samples, of which 220 are open-sourced.\\
\textsuperscript{C} Benchmark imposes a fixed time limit per task and evaluates human outputs based on work completed within that time-frame and allows the use of LLMs.}

\end{threeparttable}
\label{tab:existing_benchmarks}
\end{table*}
\endgroup

FrontierFinance provides several distinct advantages over current evaluation frameworks:
\begin{itemize}[leftmargin=2em]
    \item \textbf{End-to-end workflow}: Tasks are grounded in real industry workflows and developed by domain experts to reflect the full pipeline, from data sourcing to final deliverables, while incorporating practical considerations such as time and cost.
    
    \item \textbf{Computer-use}: Tasks span multiple data formats and tools, requiring agents to identify, retrieve, and extract relevant information from diverse sources.
    
    \item \textbf{Rubric-based evaluation}: Each task is accompanied by a detailed rubric tailored to its model type, with 5 to 11 evaluation categories and up to 140 scoring criteria. Because financial model types differ substantially in structure and mechanics, the rubrics define model specific sub-categories with graduated scoring tiers to enable consistent and reproducible evaluation across tasks.
    
    \item \textbf{Long-horizon}: Tasks are inherently long-form, requiring an average of over 18 hours of expert effort to complete, capturing the sustained reasoning and execution needed in real-world financial modeling.

    \item \textbf{Reference answers}: Each task is accompanied by a reference finance model, produced and vetted by domain experts.
\end{itemize}

\section{Benchmark Construction}

\subsection{Dataset Design and Construction}

FrontierFinance is designed to evaluate the ability of LLMs to perform long-horizon, end-to-end financial modeling tasks. Each task requires an agent to construct a complete financial model from scratch given a detailed prompt and evaluation rubric. This includes identifying and incorporating relevant financial data, forming appropriate assumptions, performing multi-step calculations, and generating forward-looking projections, closely reflecting the workflow of professional financial analysts. \Cref{fig:lbo_task_sample} shows an excerpt of a sample LBO task.

The benchmark spans five core financial model types, with five tasks per category, for a total of 25 tasks. A detailed description of each model type is provided in \Cref{app:finance_background}. These tasks are designed to capture a diverse set of real-world scenarios across industries, capital structures, and analytical objectives. Within each model type, the dataset includes five distinct company-specific questions, ensuring variation in both financial context and modeling complexity.

The dataset was developed from the ground up in collaboration with a former investment banker and an AI data expert, informed by real-world financial modeling practices and public company disclosures. Each model type required approximately 27.5 hours to design and construct, reflecting the level of detail needed to ensure realism and consistency. Tasks are structured to capture practical analytical requirements, incorporating relevant financial context, constraints, and decision-making objectives. To mitigate potential memorization bias in LLM training data, tasks are based on recent financial information and designed to require synthesis and reasoning rather than simple recall.

\begingroup
\setlength{\tabcolsep}{4pt}

\begin{table*}[t]
\centering
\small
\resizebox{0.9\textwidth}{!}{%
\begin{tabular}{l c c c c c l}
\toprule

\makecell{\textbf{Finance} \\ \textbf{Model}} &
\makecell{\textbf{Difficulty} \\ \textbf{(10)}} &
\makecell{\textbf{Estimated} \\ \textbf{Human Hours}} &
\makecell{\textbf{Min. \# Docs} \\ \textbf{Referenced}} &
\makecell{\textbf{\# Unique} \\ \textbf{Sources}} &
\makecell{\textbf{\# Rubric} \\ \textbf{Steps}} &
\textbf{Deliverables} \\

\midrule

M\&A            & 7.6 & 12 - 16 & 6 & 1 & 11 & Excel \\
Three-statement & 7.5 & 13 - 16 & 5 & 2 & 5 & Excel \\
LBO             & 7.1 & 11 - 15 & 6 & 2 & 9 & Excel, PPT \\
Lender          & 6.7 & 10 - 12 & 3 & 2 & 7 & Excel \\
DCF             & 6.6 & 8 - 10 & 6 & 4 & 9 & Excel \\

\bottomrule
\end{tabular}
}
\caption{Overview of FrontierFinance complexity across financial model types.}
\label{tab:frontiner_finance_overview}
\end{table*}

\endgroup

Following initial construction, all tasks were reviewed by a team of finance professionals to ensure clarity, correctness, and practical relevance. This validation process ensures that each problem is well-specified, internally consistent, and representative of tasks encountered in real-world financial modeling workflows.

\begin{figure}[h]
    \centering

\begin{PromptBox}{Excerpt: LBO Task Prompt; Electronic Arts}

We are evaluating a full buyout of a publicly traded business called Electronic Arts on behalf of a mega fund private equity firm. Electronic Arts develops, markets, publishes, and delivers games, content, and services across consoles, PCs, and mobile devices.
\\

\textbf{Deliverables}

\begin{itemize}[leftmargin=*]
    \item Single-sheet LBO model (Excel) with all sections specified below:
        \begin{itemize}[leftmargin=2em]
            \item Transaction Drivers
                \begin{itemize}[leftmargin=2em]
                    \item Equity Value
                    \item Enterprise Value
                    \item 5-year Returns (which links to the calculations below)
                    \item Model Assumptions
                    \item Sources and Uses
                \end{itemize}
                
            \item Financial Forecast
            \item Debt Schedule
            \item Returns Calculation
            \item Sensitivity Tables
            \item Bridge Schedule
            
        \end{itemize}

    \item{Sensitivity tables on entry premium vs. exit multiple (IRR and MoM)}
    \item{Investment memo in PPT format covering executive summary, business assessment, market analysis, investment thesis, deal terms and rationale, risks and mitigants, and key diligence items.}\\
\end{itemize}

\textbf{Data Sources}
\begin{itemize}[leftmargin=*]
    \item{Pull all historical financials directly from Electronic Arts’ SEC filings: FY2024 10-Ks / 10 Qs, FY2025 10-Ks / 10 Qs. For FY2026, use 10Qs and 8Ks from Q1 through Q3 as filed}

    \item{When the provided deal assumptions outlined in Section B conflict with public filings, the provided assumptions take precedent}

\end{itemize}

\end{PromptBox}

    \caption{An excerpt of an LBO task for Electronic Arts. This figure shows only a small portion of the full task; the complete task is provided in ~\Cref{fig:lbo_rubric_full}.}
    \label{fig:lbo_task_sample}
\end{figure}

\subsection{Rubric-Based Evaluation}
To enable systematic and scalable evaluation, each financial modeling task is accompanied by a detailed rubric designed by a finance professional with domain expertise. The rubric serves as a structured scoring framework that captures multiple dimensions of model quality. The total achievable score varies across model types, reflecting differences in complexity and the level of effort required to construct each type of financial model.

Each rubric incorporates an initial gating condition to filter out fundamentally invalid submissions. This condition is intended to identify cases where the model output is unusable, such as when improper financial data sources are used, when the generated model pulls data from an incorrect fiscal period, or when the output consists entirely of static values with no underlying formula logic. By applying this early filter, the evaluation process avoids unnecessary grading effort on outputs that do not meet minimum validity criteria.

Beyond this gating step, each rubric is organized into 5--11 evaluation categories depending on the model type, each category corresponding to a key aspect of financial modeling quality (e.g., data extraction, financial assumption validity, logical consistency, and structure). Within each category, multiple scoring levels are defined with precise descriptions to ensure consistent and interpretable grading. The scoring scheme is designed to reflect the practical effort required to correct errors: mistakes that would require substantial human intervention are penalized more heavily than minor issues that can be easily fixed.

The level of detail in the rubric is intentionally high to support automated evaluation using an LLM-as-a-judge framework. By providing explicit scoring guidelines and well-defined criteria, the rubric reduces ambiguity in grading and enables a reliable approximation of expert human judgment, reducing the manual effort required to evaluate models at scale. The rubric used for the LBO task is presented as an example in \Cref{fig:lbo_rubric_detailed}.

\section{Experiments}

\subsection{Agent Setup and Evaluation Protocol}

We evaluated FrontierFinance using a set of state-of-the-art LLM-based agents, including GPT-5.4, Opus 4.6, Sonnet 4.6, and Gemini 3.1 Pro, that support computer use. Harness design is a critical factor influencing model performance \citep{lee2026metaharnessendtoendoptimizationmodel}, however, our goal was to provide a set of simple and reproducible LLM baselines. Accordingly, we prioritize simplicity in our setup.

Each model is provided with a minimal harness consisting of (i) the ability to control a GUI via its respective computer-use API and (ii) access to a bash terminal. The full system prompt is included in \Cref{fig:system_prompt}. Additional details on model configurations are provided in \Cref{tab:model-config}. We leave exploration of more complex agentic designs and harness optimizations to future work. We implement all of our harnesses using Inspect \citep{UK_AI_Security_Institute_Inspect_AI_Framework_2024}.

\subsection{Human Evaluation of Generated Models}

Due to the significant demands of professional financial evaluation, averaging over 2.5 hours per task, we conducted a manual evaluation on a representative subset of 10 modeling tasks. While numerically small, this subset constitutes over 180 hours for an expert to construct financial models from scratch (\Cref{app:modeling_team} describes the modeling process and guidelines) and over 50 hours of additional expert labor to grade. The tasks were sampled to capture variation in complexity within different finance model types. We evaluate the two highest-performing agents from our evaluation suite--GPT-5.4 and Opus 4.6--and models produced by human experts. Each model was graded in accordance with the provided rubric by domain experts. Details about the grading process and annotation agreement is provided in \Cref{app:grading_guidelines}.

\subsection{Rubric Guided LLM Judge}

To enable scalable evaluation, we explore the use of an LLM-based judge. The judge is provided with expert-defined rubrics and instructed to assess the models produced by the LLMs for each task. We reuse the same harness as in the evaluation stage, but disable computer-use capabilities during judging. We use Sonnet 4.6 as our underlying judgment agent due to its balance between performance and overall cost. We conduct LLM-based evaluation for all agents in our suite across the full set of 25 benchmark tasks.

To facilitate efficient and accurate evaluation, the judge is equipped with two tools: \texttt{edgar}, which enables querying the SEC EDGAR\footnote{Available at \url{https://www.sec.gov/search-filings}} database, and \texttt{calc}, a custom tool which supports reading and analyzing Excel files.

We consider two variants of the judge: one with access to detailed rubrics and one without. In the rubric-free setting, the judge is provided with a coarse scoring guideline (on a 0–100 scale) written by someone without financial expertise. The full prompt is available in \Cref{ref:rubric_analysis}. In the rubric-guided setting, the judge is given the relevant rubric sections in a structured, category-by-category manner. This setup allows us to evaluate the impact of rubric specificity on judgment quality and consistency.

\begin{table}[t]
\centering
\resizebox{\textwidth}{!}{%
\begin{tabular}{llccc|ccc}
\toprule

\multirow{2}{*}{\makecell{\textbf{Finance} \\ \textbf{Model Type}}}
& \multirow{2}{*}{\textbf{Task}}
& \multicolumn{3}{c}{\textbf{Score (\%)}}
& \multicolumn{3}{c}{\textbf{Time Taken (h)}} \\

& & \textbf{Human Expert} & \textbf{GPT 5.4} & \textbf{Opus 4.6} & \textbf{Human Expert} & \textbf{GPT 5.4} & \textbf{Opus 4.6}\\

\midrule

\multirow{2}{*}{M\&A}
 & Cisco-Splunk & {\bf 77.5} & 69.9 & 72.9 & 20 & 0.73 & 0.63 \\
& JnJ-Shockwave & 73.0 & {\bf 80.6} & 75.6 & 17 & 0.75 & 0.70 \\
\midrule

\multirow{2}{*}{3 Statement}
& Merck \& Co. & \highlight{67.6} & {\bf 73.8} & \highlight{\textcolor{rust}{\textbf{36.5}}} & 35 & 0.80 & 0.51 \\
   & Rackspace & {\bf 77.4 } & \highlight{49.2} & \highlight{67.6} & 35 & 1.01 & 0.81 \\
\midrule

\multirow{2}{*}{LBO}
& Electronic Arts & 72.2 & {\bf 77.1} & 76.1 & 19 & 0.77 & 0.86 \\
       & Walgreens & {\bf 74.4} & \highlight{61.6} & \highlight{\textcolor{rust}{\textbf{35.3}}} & 21 & 1.27 & 0.65 \\
\midrule

\multirow{2}{*}{Lender}
& Carnival Co. & {\bf 79.8} & 75.8 & 71.5 & 7.5 & 0.66 & 0.57 \\
         & MGM & {\bf 76.0} & 71.6 & \highlight{62.5} & 6.5 & 0.84 & 1.26 \\
\midrule

\multirow{2}{*}{DCF}
& Autozone & {\bf 88.6} & 78.7 & \highlight{58.7} & 11 & 1.52 & 0.59 \\
    & Thor & { \bf 85.0 } & 70.5 & \highlight{61.1} & 11 & 1.32 & 0.50 \\

\bottomrule \addlinespace[2pt]
& \textbf{Average} & \textbf{77.2} & \textbf{70.9} & \textbf{61.8} & \textbf{18.3} & \textbf{0.97} & \textbf{0.71} \\
\bottomrule
\end{tabular}%
}
\caption{Expert-graded performance comparison of human experts and AI agents on financial modeling tasks. Cells highlighted in yellow indicate cases where revising the generated model is less efficient than rebuilding it from scratch. Scores in \textcolor{rust}{red} indicate the model has failed the gating condition.}

\label{tab:human_graded_results}
\end{table}

\section{Results}

\subsection{Human Evaluated Models}
\Cref{tab:human_graded_results} presents an expert-graded comparison of human and LLM agent performance on financial modeling tasks. Overall, human experts outperform both GPT-5.4 and Opus 4.6, achieving an average score of 77.2\%, compared to 70.9\% and 61.8\%, respectively. While LLM-generated outputs are often directionally correct, they frequently lack the consistency and completeness required for professional use.

In terms of efficiency, LLM agents complete tasks significantly faster than human experts. On average, GPT-5.4 and Opus 4.6 require 0.97 and 0.71 hours per task, respectively, compared to 18.3 hours for human experts (see \Cref{sec:time_diff} for a discussion on time estimates). However, this speed advantage comes with substantial variability in output quality. 

To better characterize output quality, we distinguish---based on human grader reviews---between models that can be reasonably refined and those that require rebuilding from scratch. Human experts produced only one model that would require a full restart, compared to two for GPT-5.4 and six for Opus 4.6. This reinforces a key limitation of current LLM agents: even when partially correct, their outputs often fail in ways that make iterative correction inefficient relative to reconstruction.

The open-ended nature of the tasks led to differing computer-use strategies across agents. GPT-5.4 primarily used computer-use for information retrieval, whereas both Opus 4.6 and Sonnet 4.6 visually inspected their spreadsheets prior to submission. \Cref{fig:fig_test} illustrates examples of these behaviors, and \Cref{fig:tool_use_extended_appendix} presents patterns across all 10 samples graded by experts. Notably, Gemini 3.1 Pro did not use the computer-use tool at all.

\subsection{LLM Judge Evaluated Models}

\begin{figure}[t]
    \centering
    \begin{minipage}[t]{0.485\textwidth}
        \centering
        \includegraphics[width=\linewidth]{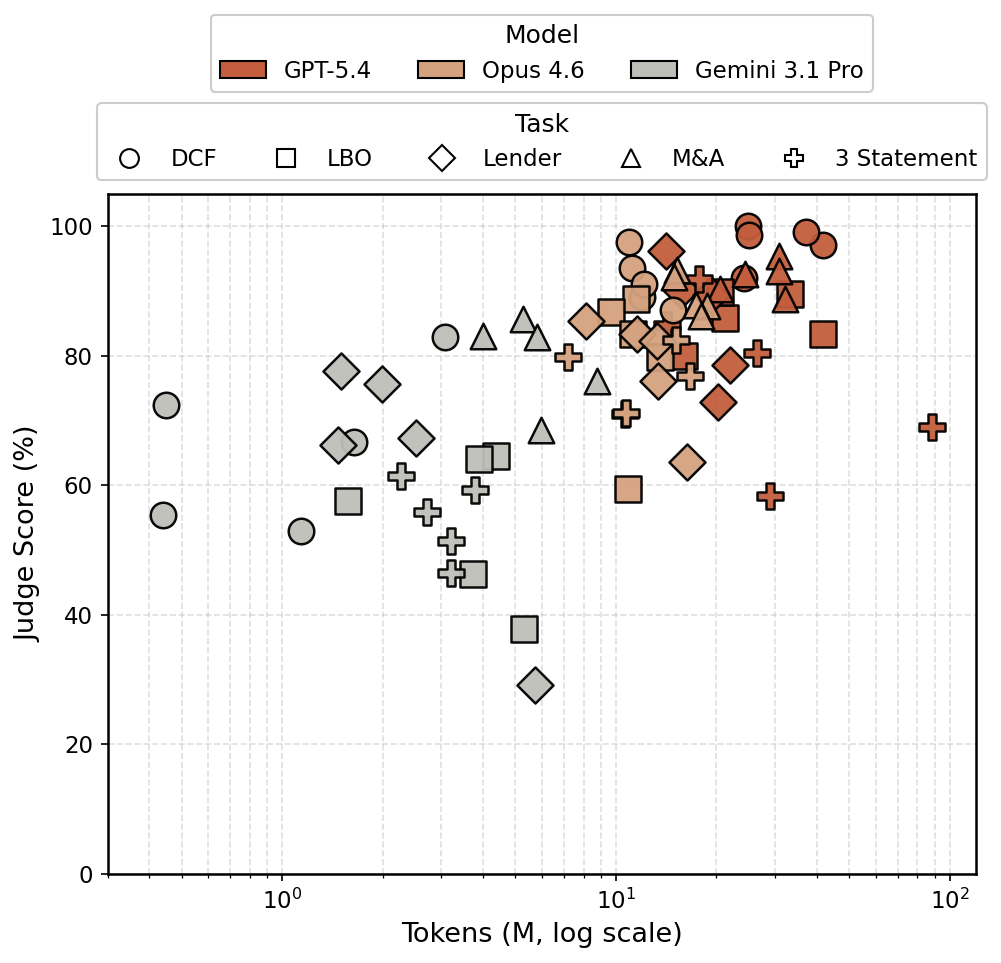}
        \captionof{figure}{Analysis of token consumption versus overall score across finance model types and LLM agents, excluding the LLM Judge Sonnet 4.6.}
        \label{fig:token_performance}
    \end{minipage}
    \hfill
    \begin{minipage}[t]{0.485\textwidth}
        \centering
        \includegraphics[width=\linewidth]{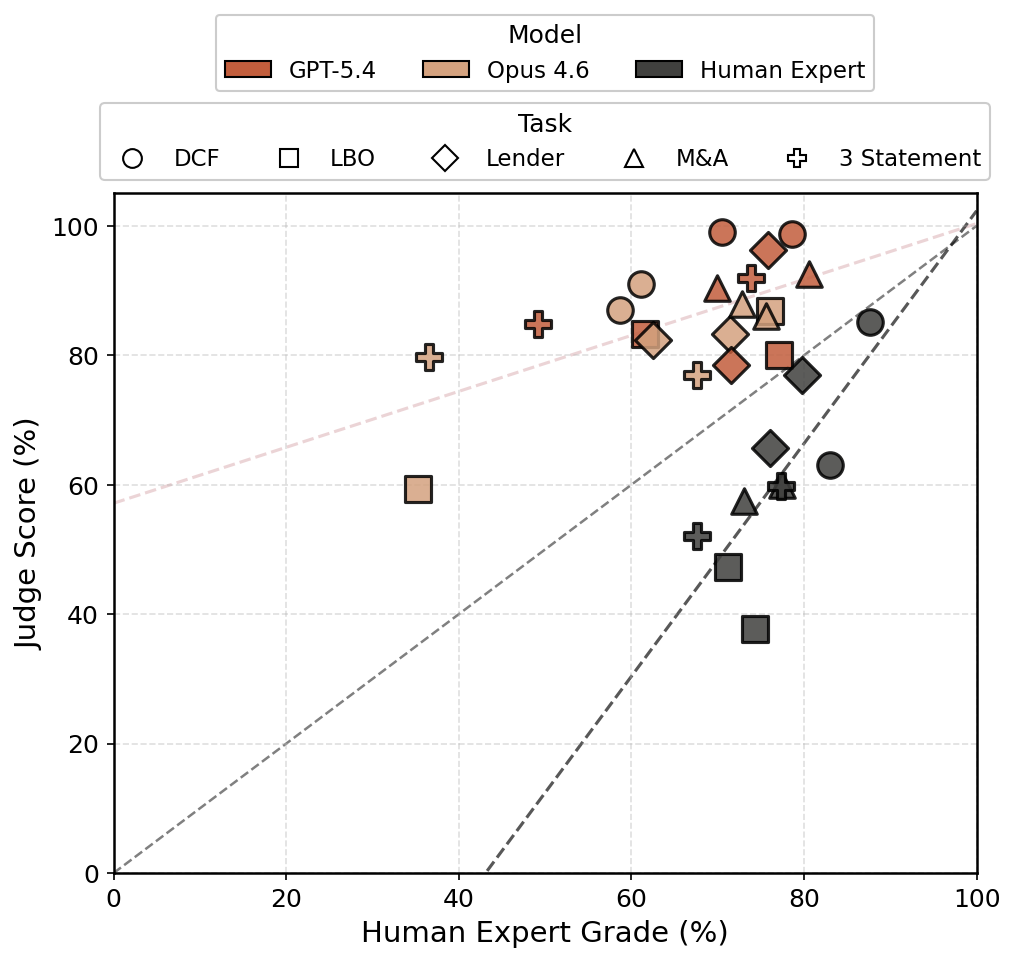}
        \captionof{figure}{Performance of the Rubric-Guided LLM Judge against Human Expert grades across all finance model types.}
        \label{fig:judge_vs_human}
    \end{minipage}
\end{figure}

Among the four agents evaluated by an LLM Judge, as shown in \Cref{tab:model-stats}, GPT-5.4 processes substantially more tokens per task (27.6M total) than Opus 4.6 (13.0M), and achieves higher performance. Meanwhile, Gemini 3.1 Pro operates at a much lower token budget (3.1M), but also exhibits significantly reduced capability. This trend is also observed in \Cref{fig:token_performance}. Sonnet 4.6 is excluded from this analysis due to evidence of significant self-evaluation bias in its corresponding judge. However, \Cref{tab:detailed_perf} reports the average judge scores across all 25 tasks for the full evaluation suite.

\Cref{fig:judge_vs_human} visualizes the performance of the LLM judge framework against human expert grades across all finance model types. We observe that, even when provided with a detailed rubric, the LLM judge consistently overestimates performance of LLM-generated outputs relative to human graders. Based on experiments conducted with and without rubric guidance, we find that incorporating rubrics substantially improves correlation with human scores, increasing it from 0.20 to 0.62. Further details are provided in \Cref{ref:rubric_analysis}.

\section{Analysis \& Discussion}

\subsection{LLM performance}
Manual evaluation of outputs from both GPT-5.4 and Opus 4.6 against the rubric reveals a clear pattern in strengths and limitations. Across tasks, both agents perform well in the initial “intake” phase: they are effective at extracting relevant information from source documents, synthesizing qualitative insights, and generating coherent high-level rationales. They also offer substantial speed advantages, producing complete model drafts approximately 20× faster than human analysts (\Cref{tab:human_graded_results}).

However, this competence does not extend to the construction of robust, auditable financial models. Both agents consistently fail to translate extracted insights into precise, structured outputs that meet professional standards. Several recurring failure modes emerge:

\textbf{State tracking and consistency.}
The tasks require maintaining a coherent representation of how information flows through the model. Both agents struggled with this. They frequently generated incorrect references, such as combining unrelated quantities within a single formula or linking projections to the wrong time periods. 
In settings with interdependent calculations, such as feedback loops, neither agent was able to correctly maintain the dependency structure, often resorting to incorrect or incomplete implementations.

\textbf{Program synthesis and execution.}
A central limitation was the inability to construct outputs through executable logic. Both agents often replaced formulas with static values, producing models that appeared complete but could not be updated or graded. 
When faced with more complex tasks, the agents avoided standard solutions and instead approximated outputs through ad-hoc workarounds. For example, an agent built roughly 88 hidden rows with missing labels formatted in white font to replicate what a single built-in data table formula does natively, concealing the workaround behind two layers of obfuscation.

\textbf{Instruction following and domain alignment.}
Both agents inconsistently followed instructions, completing some requirements while omitting others. Key elements such as required checks, organization, or calculation methods were often missing. Outputs also deviated from widely accepted structural and formatting conventions, making them harder to interpret and review. These issues go beyond presentation: they reflect a lack of alignment with the implicit standards that govern how such tasks are typically completed.

\textbf{Grounding and verification.}
The outputs occasionally contained unsupported or fabricated values embedded within otherwise valid results, making errors difficult to detect without detailed inspection. More broadly, neither agent demonstrated meaningful self-verification, such as basic consistency checks. In several cases, even when validation mechanisms were included, they were either bypassed or implemented in ways that rendered them ineffective. For example, balance sheets were often balanced with implausible, fabricated values merely to satisfy the balancing criteria specified.

\begin{table}[t]
\centering

\label{tab:model-stats}
\resizebox{\textwidth}{!}{%

\begin{tabular}{l|c|cccc|cccc}
\toprule
\multirow{2}{*}{\textbf{Model}} & \textbf{Average} & \multicolumn{4}{c|}{\textbf{Average Tokens}} & \textbf{Working} & \textbf{Total }  & \textbf{Computer} & \textbf{Cost} \\

 & \textbf{Score (\%)} & \textbf{Total} & \textbf{Input} & \textbf{Output} & \textbf{Reasoning} & \textbf{Time (h)} & \textbf{Messages}  & \textbf{Use \%} & \textbf{(\$)}\\
 
\midrule
  GPT-5.4 & 87.5  & 27.6M & 27.4M & 182k & 89k & 1.0 & 302 & 8\% & 14.84 \\
  Opus 4.6 & 83.0 & 13.0M & 12.9M & 128k & 3k & 0.7 & 196 & 24\% & 16.00 \\
  Sonnet 4.6 & 84.0 & 24.9M & 24.7M & 209k & 23k & 1.3 & 274 & 23\% & 13.69\\
  Gemini 3.1 Pro & 63.5 & 3.1M & 3.1M & 57k & 14k & 0.3 & 122 & 0\% & 2.12 \\  
\bottomrule
\end{tabular}
}
\caption{Average token usage, time, and tool statistics across LLM agents evaluated per task. Average score is computed by the Sonnet 4.6-as-a-Judge over the full dataset. We provide a more detailed performance breakdown in \Cref{tab:detailed_perf}.}

\end{table}

\subsection{Human Performance}

We observe that human experts consistently produce outputs that align with professional standards in both structure and execution. Their models are well-organized and are mechanically sound. Formulas are correctly linked, outputs are auditable, and model structures are intuitive for downstream review. Core functionalities---such as sensitivity tables, circularity toggles, and integrated assumption flows---are implemented correctly, ensuring that models are dynamic, traceable, and usable in practice. Compared to LLM agents, humans also demonstrate stronger adherence to task instructions and deliverables.

However, limitations emerge primarily in areas of judgment and analytical rigor. While models are functionally correct, forecasting decisions are occasionally simplistic or insufficiently justified. Common issues include reliance on heuristic approaches (e.g., flat percentage-of-sales assumptions), use of short historical averages despite visible trends, and application of static growth rates without clear rationale. These choices do not break the model mechanically but can weaken the quality and credibility of the analysis. There is a $\sim$30\% difference in the estimated time taken to complete the task and the actual time taken. We discuss this discrepancy in detail in \Cref{sec:time_diff}.

\section{Related Work}

\emph{\textbf{LLM Benchmarks}}
A large body of work has evaluated the reasoning capabilities of large language models using standardized benchmarks. Datasets such as MMLU \citep{hendrycks2021measuringmassivemultitasklanguage} and BIG-bench \citep{srivastava2023imitationgamequantifyingextrapolating} measure broad knowledge and multi-task reasoning, while more challenging benchmarks such as GPQA \citep{rein2023gpqagraduatelevelgoogleproofqa} focus on graduate-level, domain-specific questions designed to resist memorization.

Building on these general reasoning capabilities, domain-specific benchmarks have emerged to evaluate numerical and logical proficiency within the financial domain. Early efforts such as FinQA \citep{chen-etal-2021-finqa} and ConvFinQA \citep{chen-etal-2022-convfinqa} established standards for multi-step arithmetic reasoning over financial reports, while TAT-QA \citep{zhu-etal-2021-tat} introduced challenges requiring reasoning across hybrid tabular and textual data. Meanwhile BizBench \citep{krumdick-etal-2024-bizbench}, FinanceBench \citep{islam2023financebenchnewbenchmarkfinancial}, and XFinBench \citep{zhang-etal-2025-xfinbench} evaluate models on intricate business logic and expert-level financial problem-solving. In parallel, \cite{thorne2025largelanguagemodelsspreadsheets} evaluate LLM proficiency in spreadsheet-specific formula generation and auditing. However, these benchmarks primarily focus on discrete operations rather than end-to-end construction of interconnected financial models.

\emph{\textbf{Long-Horizon Reasoning and Tool-Use}}
Recent work has increasingly focused on evaluating LLMs in long-horizon settings that require sustained reasoning and tool integration. SWE-Bench \citep{jimenez2024swebenchlanguagemodelsresolve} and SWE-Lancer \citep{miserendino2025swelancerfrontierllmsearn} evaluate repository-level software engineering tasks, requiring models to navigate large codebases and produce executable fixes. Additional code-centric agentic benchmarks include HCAST \citep{rein2025hcasthumancalibratedautonomysoftware}, MLE-Bench \citep{chan2025mlebenchevaluatingmachinelearning}, and LongCLI-Bench \citep{feng2026longclibenchpreliminarybenchmarkstudy}, which further stress multi-step reasoning and execution in complex environments.

Beyond code generation, APEX-Agents \citep{vidgen2026apexagents} and MCP-RADAR \citep{gao2025mcpradarmultidimensionalbenchmarkevaluating} introduce real-world, multi-step tasks that combine reasoning with external tool use, highlighting the challenges of maintaining coherence over extended action sequences. Similarly, GDPval \citep{patwardhan2025gdpvalevaluatingaimodel} spans multiple domains and requires outputs in diverse formats, including PDFs, presentations, videos, and spreadsheets, reflecting the growing emphasis on evaluating end-to-end task completion.

\emph{\textbf{LLM-as-a-judge and Rubric-based evaluation}}
The use of LLMs as automated evaluators, or LLM-as-a-judge, has emerged as a scalable alternative to prohibitive human annotation costs for open-ended tasks. \citet{zheng2023judgingllmasajudgemtbenchchatbot} formalized this approach with MT-Bench \citep{Bai_2024} and Chatbot Arena \citep{chiang2024chatbotarenaopenplatform}, demonstrating that frontier models can achieve high alignment with human experts when provided with clear grading instructions. Similarly, G-Eval \citep{liu-etal-2023-g} utilizes chain-of-thought prompting and formulates evaluation as a multi-criteria scoring task. Recent work has also explored the use of Prometheus \citep{kim2024prometheusinducingfinegrainedevaluation}, an open-source evaluator model trained specifically to follow complex rubrics, highlighting the move toward specialized judging architectures. 

That being said, professional domains like finance require evaluation grounded in verifiable domain logic rather than relative preference. Consequently, recent agentic benchmarks have shifted toward deterministic rubric and rule-based evaluation. \textbf{WebArena} \citep{zhou2024webarenarealisticwebenvironment} and \textbf{OSWorld} \citep{xie2024osworldbenchmarkingmultimodalagents} utilize functional verification to provide deterministic success metrics for multi-step agents. This paradigm is further exemplified by SWE-bench \citep{jimenez2024swebenchlanguagemodelsresolve}, which relies on the execution of unit tests to verify the correctness of software engineering solutions. 

\section{Conclusion}

We introduce FrontierFinance, a novel benchmark designed to evaluate LLM agents on the realistic, long-horizon financial modeling tasks that reflect professional workflows. Through structured evaluation with granular, expert-defined rubrics, we show that while frontier LLM agents demonstrate strong capabilities in data extraction and intermediate reasoning, they fall short in producing reliable, auditable, and client-ready outputs. Our analysis reveals that even a modest performance gap (such as 6\%) relative to human experts represents a critical threshold in professional utility; such discrepancies often manifest as systemic failures in model integrity and auditability. These findings suggest that while the speed of LLM agents is transformative, their current inability to maintain structural consistency over extended horizons remains a significant barrier to the full automation of cognitively demanding professional roles.

\section*{Ethics Statement}
There is a risk that FrontierFinance results may be misinterpreted as evidence that LLMs can replace financial professionals. This work is intended solely to assess the current capabilities of LLMs on complex, real-world long-horizon tasks and to contribute to ongoing discussions around their potential impact. We do not claim that these systems are capable of fully substituting human expertise, and our results highlight the continued need for human oversight and validation.

All company data used in this benchmark is sourced from publicly available materials. However, the tasks incorporate additional assumptions and hypothetical scenarios designed to reflect possible realistic financial workflows. These assumptions are illustrative in nature and should not be interpreted as actual financial analysis or investment advice.

\section*{LLM Use Disclosure}
Claude code was used in the development of the underlying evaluation, and to help create plots.

\bibliographystyle{colm2026_conference}

\appendix
\crefalias{section}{appendix}
\section{Finance Background}
\label{app:finance_background}
FrontierFinance focuses on five core financial model types that form the analytical backbone of corporate finance, used routinely across investment banking, private equity, and credit analysis. These five core models are built in live transactions to price acquisitions, underwrite credit, structure leveraged buyouts, and value companies. They require integrating structured data from regulatory filings, layering in forward-looking assumptions, and executing multi-step calculations over extended projection horizons where errors in any one component can cascade through the entire output.

\subsection{Three-Statement Model}
A three-statement model links the income statement, balance sheet, and cash flow statement into a single, coherent framework. It is used to forecast a company’s financial performance over time by projecting revenues, expenses, assets, liabilities, and cash flows. The defining characteristic of this model is that all three statements must remain internally consistent: changes in one statement propagate through to the others. Building such a model requires careful handling of accounting principles, such as how net income flows into retained earnings and how cash flow reconciles across operating, investing, and financing activities.

\subsection{Leveraged Buyout (LBO) Model}
An LBO model analyzes the acquisition of a company using a significant amount of debt financing. The model evaluates whether the target company’s future cash flows are sufficient to service and repay the debt while generating attractive returns for equity investors. Key components include projecting operating performance, constructing a debt schedule, and estimating investor returns (e.g., internal rate of return, multiple of money, etc.). LBO models are widely used in private equity and emphasize the interaction between leverage, cash flow generation, and exit assumptions.

\subsection{Discounted Cash Flow (DCF) Model}
A DCF model estimates the intrinsic value of a business by forecasting its future free cash flows and discounting them to present value using a discount rate that reflects risk. The model typically includes explicit projections over a finite horizon and a terminal value to capture long-term growth. Central elements include assumptions about revenue growth, margins, capital expenditures, and the cost of capital. DCF analysis is a standard valuation method and requires both financial reasoning and careful handling of long-term assumptions.

\subsection{Merger Model}
A merger model evaluates the financial consequences of combining two companies into a single entity. It is commonly used to assess whether a transaction is accretive or dilutive to key metrics such as earnings per share. The model incorporates the financials of both companies, transaction structure (cash, stock, or hybrid), potential synergies, and deal-related adjustments. This type of model requires reasoning about how multiple entities interact post-transaction and how changes in capital structure and operations affect overall performance.

\subsection{Lender Model}
A lender model focuses on assessing a company’s ability to meet its debt obligations. It is used in credit analysis to evaluate repayment capacity, risk, and covenant compliance. The model typically projects cash flows and calculates financial ratios such as interest coverage and leverage. Stress scenarios are often incorporated to test how adverse conditions affect the borrower’s ability to service debt. This model emphasizes downside risk analysis and the stability of cash flows under varying assumptions.

\section{Finance Modeling Process and Guidelines}
\label{app:modeling_team}

Out of the 25 tasks, 10 were completed by finance professionals with 2--5 years of overall experience, including 2--3 years of core financial modeling experience. These experts were provided with the same instructions as the LLM agents---namely, the task definition and the evaluation rubric---to ensure consistency in expectations.

To enable a fair comparison, participants were asked not to use plug-ins or external automation tools that they would typically rely on in practice. This constraint was intended to mirror the capabilities of the evaluated LLM agents, which were not given access to such tools. The participants were not involved in the construction of the benchmark and had no prior exposure to the tasks.

All models were produced independently as part of this study. The participating professionals contributed their time voluntarily and were not financially compensated.

\begin{figure}
    \centering
     \includegraphics[width=\textwidth]{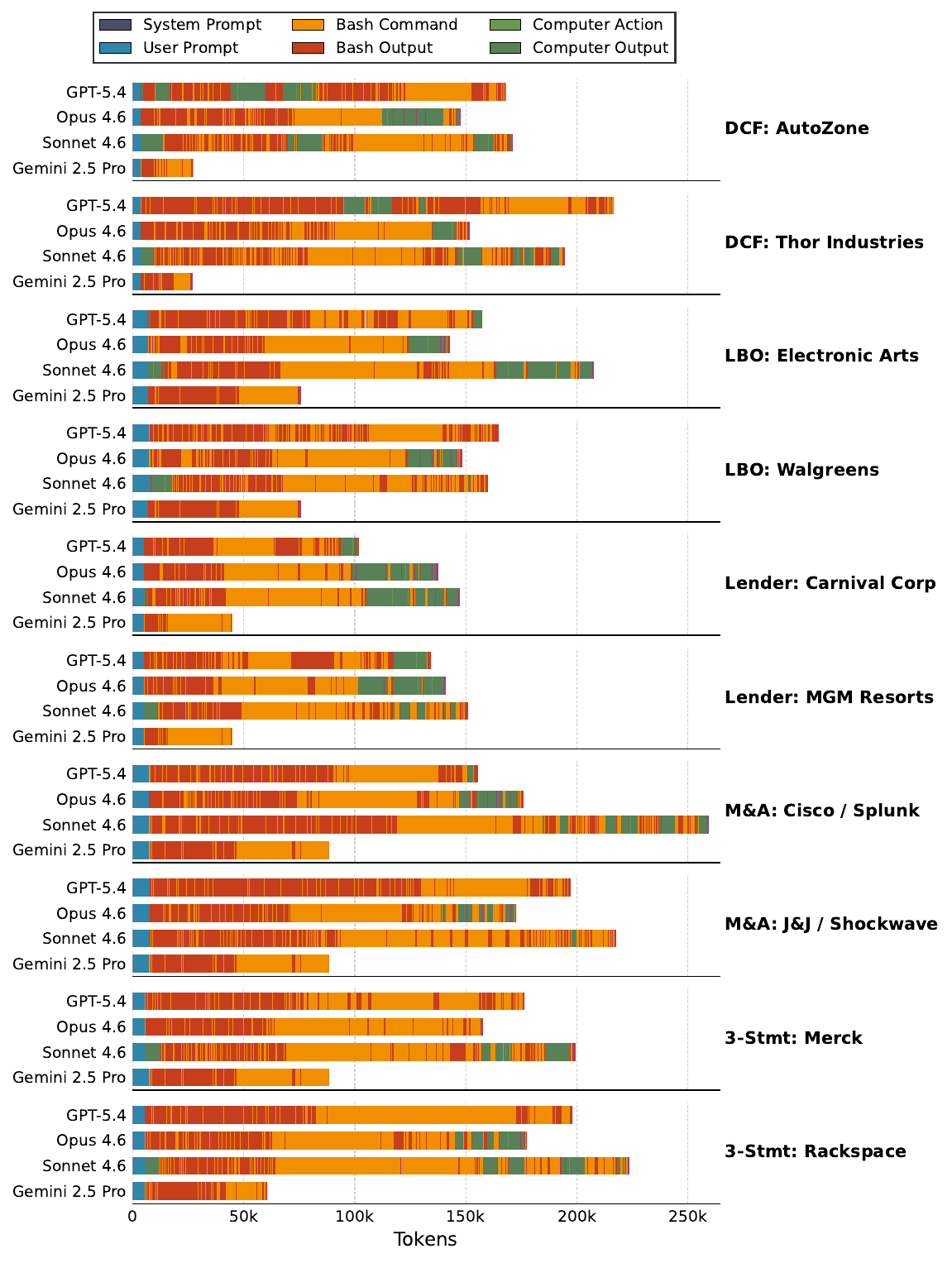}
     \caption{Tool and token use distribution visualization.}
     \label{fig:tool_use_extended_appendix}
\end{figure}

\section{Grading Process and Guidelines}
\label{app:grading_guidelines}
Out of the 25 tasks, 10 tasks (2 per finance model type) were graded by humans for models generated by humans, Opus, and GPT. Human-generated models were evaluated by a finance expert, while LLM-generated models were graded by a team of trained financial data annotators. Although the annotators were familiar with financial models (with approximately 2–3 years of experience working with financial data and degrees in finance) their expertise was primarily in building templates rather than constructing financial models from scratch. The annotation process was overseen by a finance analyst, who provided guidance and verified the scorecards post grading. The annotation effort was conducted by colleagues as part of this research and was not a separately compensated task.

Each finance model type was accompanied by a detailed rubric to ensure consistent evaluation. Each model was graded by multiple annotators: two annotators per model for Merger, Lender, and DCF, and three annotators per model for Three-Statement and LBO models. Annotators were instructed to evaluate each model holistically, considering the full structure and outputs in the context of the detailed task requirements.

\subsection*{Annotation Agreement}
We evaluate the reliability of our manual audit by calculating the Pearson correlation between the primary graders and the lead expert. We observe a high scoring correlation of 0.965 between graders and the expert, with an inter-annotator correlation of 0.986 among the graders themselves, indicating a high level of consistency in applying the rubric. 

221Scores between the expert and graders largely align across major categories such as data retrieval and source compliance. Disagreements rarely concern numerical correctness; rather, they center on whether models are constructed in a manner that is verifiable, maintainable, and reliable.

\section{Rubric Analysis}
\label{ref:rubric_analysis}

\begin{figure}
    \centering
     \includegraphics[width=\textwidth]{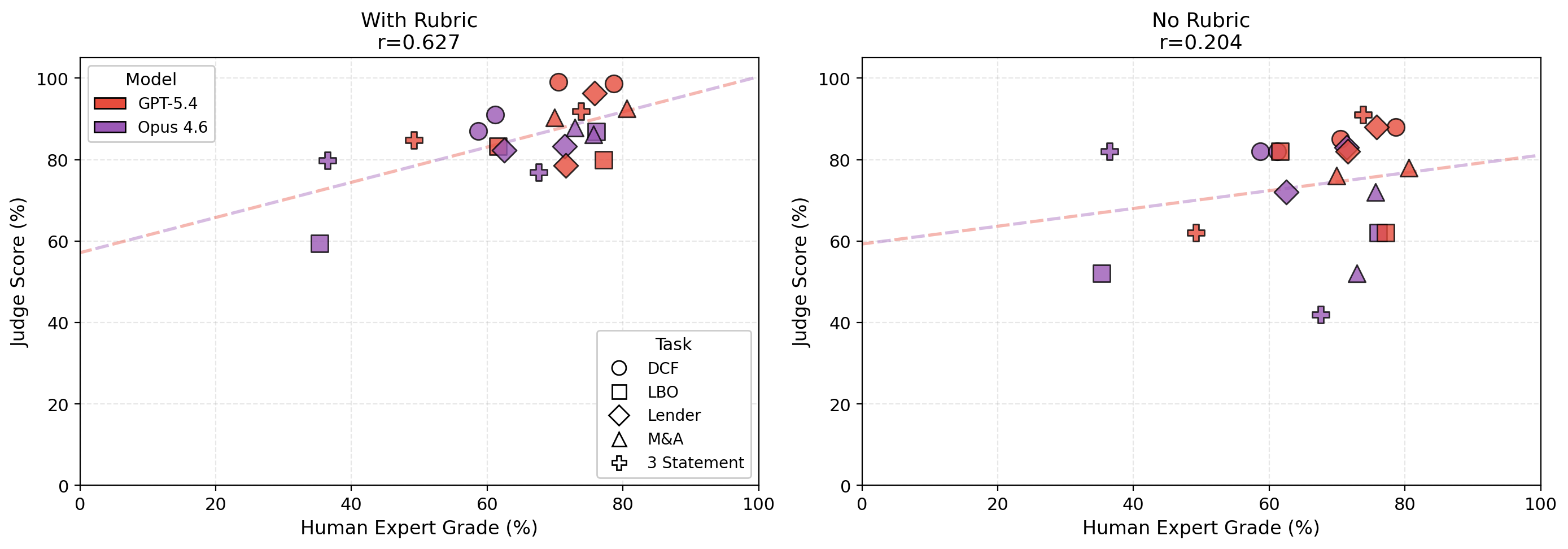}
     \caption{Comparison of the LLM Judge performance with and without the rubric}
     \label{fig:rubric_works}
\end{figure}

To validate the use of the rubric, we compare the performance of the LLM Judge with and without access to it. As illustrated in \Cref{fig:rubric_works}, the introduction of a structured rubric significantly enhances the reliability of the LLM judge, increasing the correlation with human expert grades from a weak $r=0.204$ to a much more robust $r=0.627$.

To elicit quality scores, we used the prompt, defined in \Cref{fig:system_prompt_judge}, written by someone without financial expertise.

\begin{PromptBoxBreak}{System Prompt}
Examine the model output files carefully. Open and inspect the Excel workbook(s) to evaluate the quality of the financial model. 

Provide a single \textbf{quality score from 0 to 100}, where:
\begin{itemize}
    \item \textbf{100} = Client-ready. Could be sent to a client or used in a live deal with minimal edits. Accurate data, correct formulas, clean formatting, comprehensive analysis.
    \item \textbf{80-99} = Strong work. Minor issues but fundamentally sound and usable.
    \item \textbf{60-79} = Acceptable. Core structure is there but has notable gaps, errors, or missing sections.
    \item \textbf{40-59} = Below expectations. Significant issues — wrong data, broken formulas, missing key components. 
    \item \textbf{20-39} = Poor. Major structural problems, largely incomplete or incorrect.
    \item \textbf{0-19} = Total failure. Empty, nonsensical, or completely wrong.
\end{itemize}
 
Consider:                                                                         
\begin{enumerate}
    \item \textbf{Data accuracy} — Are historical financials correctly sourced from SEC filings? 
    \item \textbf{Model completeness} — Are all requested sections and deliverables present?  
    \item \textbf{Formula integrity} — Do formulas work correctly? Are cross-references intact?   
    \item \textbf{Formatting \& presentation} — Is the model clean, organized, and professional?\\
\end{enumerate}
                                             
After your investigation, respond with ONLY a JSON object in this exact format, then write [DONE]:\\

{"score": <0-100>, "rationale": "<2-3 sentence explanation>"}\\                  

[DONE]      

\end{PromptBoxBreak}
\captionof{figure}{LLM-as-a-judge system prompt for non-rubric baseline evaluations}
\label{fig:system_prompt_judge}

\begin{table}[h]
  \centering
  \caption{Average judge score (\%) by model and task type.}
  \label{tab:scores-by-task}
  \begin{tabular}{lcccccc}
  \toprule
  Model & DCF & 3-Stmt & LBO & M\&A & Lender & Avg \\
  \midrule
  GPT-5.4 & 97.4 & 76.9 & 85.7 & 92.0 & 85.5 & 87.5 \\
  Opus 4.6 & 91.7 & 76.2 & 79.6 & 89.4 & 78.1 & 83.0 \\
  Sonnet 4.6 & 87.1 & 80.9 & 79.4 & 86.3 & 86.5 & 84.0 \\
  Gemini 3.1 Pro & 66.1 & 54.9 & 54.1 & 79.2 & 63.2 & 63.5 \\
  \bottomrule
  \end{tabular}
  \label{tab:detailed_perf}
\end{table}

\section{Difference Between Estimated and Actual Completion Time}\label{sec:time_diff}
During dataset construction, expert estimates placed task completion time between 8--16 hours, averaging 12.5h per task (Table~\ref{tab:frontiner_finance_overview}). However, observed completion times ranged more widely, from 6.5 to 35 hours, averaging >18h, which is an under-estimate by approximately 31\%. Post-hoc analysis and discussions with participating analysts suggest the following factors driving this discrepancy:

\begin{itemize}

    \item \textbf{Idealized initial estimates:} Time estimates were provided by a finance professional, with experience in investment banking, involved in dataset and rubric design, reflecting optimized execution under familiar workflows.
    
    \item \textbf{Workflow overhead from rubrics:} Analysts were required to follow externally defined rubrics, introducing additional overhead and deviations from standard modeling workflows, which increased completion time.

    \item \textbf{Structural inconsistencies across tasks:} Observed time variation does not always align with underlying model complexity. In this work, the LBO model was constructed first, followed by the Lender model. As a result, the Lender model was completed more quickly (approximately 6.5--7.5 hours), despite sharing similar forecasting logic, projections, and statement linkages. This suggests that completion time is influenced not only by model complexity but also by familiarity and repeated exposure to similar workflows.
    
    \item \textbf{Impact of deliverables and granularity:} Differences in completion time are partly explained by non-core requirements, such as presentation outputs (e.g., PPTs) and varying levels of forecasting detail, rather than fundamental differences in modeling difficulty.

\end{itemize}

\begin{table}[t]
\centering
\resizebox{\textwidth}{!}{%
\begin{tabular}{l | l c l}
\toprule
\textbf{Model Name} & \textbf{Provider} & \textbf{Thinking Setting} & \textbf{Hyper-parameters}\\
\midrule \addlinespace[2pt]
  GPT-5.4 & OpenAI & High & \texttt{Temperature:} \texttt{default} \\[2pt]
    & & & \texttt{Reasoning Effort:} \texttt{XHigh}\\
\midrule \addlinespace[2pt]
  Opus 4.6 & Anthropic & High & \texttt{Temperature:} \texttt{default}\\[2pt]
  & & & \texttt{Reasoning Effort:} \texttt{High}\\[2pt]
  & & & \texttt{Context Window:} \texttt{1e6}\\
\midrule \addlinespace[2pt]
  Sonnet 4.6 & Anthropic & High & \texttt{Temperature:} \texttt{default}\\[2pt]
  & & & \texttt{Reasoning Effort} \texttt{High}\\[2pt]
  & & & \texttt{Context Window:} \texttt{1e6}\\
\midrule \addlinespace[2pt]
  Gemini 3.1 Pro & Google DeepMind & High & \texttt{Temperature:} \texttt{default}\\[2pt]
  & & & \texttt{Reasoning Effort:} \texttt{default}\\
\bottomrule
\end{tabular}
}
\caption{Summary of LLM agents evaluated in this study.}
\label{tab:model-config}
\end{table}

\begin{PromptBoxBreak}{System Prompt}
You are an expert investment banking associate building a financial model.

You have access to a computer with some helpful tools pre-installed (Firefox, LibreOffice, python3 + libraries). You can interact with it using:
\begin{itemize}
    \item \textbf{computer}: Control a GUI desktop
 \item \textbf{bash}: Run shell commands
\end{itemize}

IMPORTANT:
\begin{itemize}
    \item NEVER ask questions or wait for user input. Always proceed autonomously.
    \item Save all your work to /workspace/ (include README.md documenting any problems, missing data, or assumptions)
    \item When FULLY COMPLETED, respond with [DONE].
\end{itemize}

Work systematically. This is a challenging task requiring you to discover and use the right tools. We expect a human would take at least 8 hours to complete this with acceptable quality.

\end{PromptBoxBreak}
\captionof{figure}{Full system prompt for rubric-guided financial model generation for the provided task}
\label{fig:system_prompt}

\begin{PromptBoxBreak}{LBO Task Prompt; Electronic Arts}

We are evaluating a full buyout of a publicly traded business called Electronic Arts on behalf of a mega fund private equity firm. You are a VP at this megafund evaluating the full buyout of a publicly traded business called Electronic Arts. Electronic Arts develops, markets, publishes, and delivers games, content, and services across consoles, PCs, and mobile devices.
\\

\textbf{Deliverables}

\begin{itemize}
    \item Single-sheet LBO model (Excel) with all sections specified below:
        \begin{itemize}
            \item Transaction Drivers
                \begin{itemize}
                    \item Equity Value
                    \item Enterprise Value
                    \item 5-year Returns (which links to the calculations below)
                    \item Model Assumptions
                    \item Sources and Uses
                \end{itemize}
            \item Financial Forecast
            \item Debt Schedule
            \item Returns Calculation
            \item Sensitivity Tables
            \item Bridge Schedule 
        \end{itemize}
    \item{Sensitivity tables on entry premium vs. exit multiple (IRR and MoM)}
    \item{Investment memo in PPT format covering executive summary, business assessment, market analysis, investment thesis, deal terms and rationale, risks and mitigants, and key diligence items.}\\
\end{itemize}

\textbf{Data Sources}
\begin{itemize}
    \item{Pull all historical financials directly from Electronic Arts’ SEC filings: FY2024 10-Ks / 10 Qs, FY2025 10-Ks / 10 Qs. For FY2026, use 10Qs and 8Ks from Q1 through Q3 as filed}
    \item{When the provided deal assumptions outlined in Section B conflict with public filings, the provided assumptions take precedent}
\end{itemize}

\begin{enumerate}[label=(\Alph*)]
    \item \textbf{Global Formatting Rules}

    \begin{itemize}
        \item Blue font = hardcoded input. Black font = formula. Green font = cross-section link
        \item Navy fill, white bold text for all section headers
        \item Circularity toggle cell at top of sheet: 0 = interest zeroed, 1 = live iterative calc
        \item S\&U balance check cell and FCF check cell, both visible near top
        \item No embedded constants in any forecast formula. Every calculation cell traces to a blue-font assumption
    \end{itemize}

    \item \textbf{Deal Assumptions}    
    \begin{enumerate}[label=(\alph*)]
        \item \textbf{Transaction Overview}
        \begin{itemize}
            \item Announced September 29, 2025. Assume transaction close date of March 31, 2026 (FY2026 year-end)
            \item All cash take private
        \end{itemize}

        \item \textbf{Capital Structure and Transaction Assumptions}
        \begin{itemize}
            \item Assume all equity is funded by the private equity and there are no management options
            \item Assume \$300M of transaction advisory fees
            \item Assume \$300M of minimum cash
            \item Assume a fixed SOFR of 3.7\% throughout the projection period
            \item Assume transaction closes at the end of fiscal year 2026 (ended Mar 31, 2026)
            \item Prepare a dedicated FY2026 bridge schedule below all sensitivity tables that presents Q1-Q3 actuals, the Q4 seasonal projection, and the resulting FY2026 totals. The main forecasting section should not recompute these figures; instead, it should link directly to the final FY2026 totals calculated in the bridge schedule
            \begin{itemize}
                \item Q4 projection method: Calculate the Q3 to Q4 sequential (QoQ) growth rate for both FY2024 and FY2025. Average those two QoQ rates, then apply to Q3 FY2026 actuals to project Q4 FY2026, adjusted for any specific management guidance
                \item Present the build-up explicitly:
                \begin{itemize}
                    \item Q1–Q3 2026 actual results (YTD)
                    \item Q4 projected results (with seasonal basis shown)
                \end{itemize}
                \item Resulting FY2026 totals
            \end{itemize}

            \item Assume that 2026 is the last year of actuals and that April 2026 is the beginning of the forecast period
            \begin{itemize}
                \item Assume that the offer price is \$210/share
                \item \% Premium to Current Price is 25\%
            \end{itemize}

            \item FDSO: ~260.0M (249.4M common + 10.6M RSUs, options negligible)
            \item Assume all cash earn SOFR - 50bps
            \item Assume all existing net debt is refinanced
            \item LTM GAAP EBITDA as of June 30, 2025 is approximately \$2,005M. To arrive at LTM Adjusted EBITDA, add back stock-based compensation and non-recurring charges (restructuring, acquisition-related expenses, and other one-time items) per EA's SEC filings. You must build the GAAP-to-Adjusted EBITDA bridge explicitly. Use LTM Adjusted EBITDA for leverage and entry multiple calculations
            \item Base case exit multiple = entry multiple
            \item Exit Enterprise Value = (Entry EV / Entry LTM Adjusted EBITDA) × FY2031 Adjusted EBITDA
            \item Include sensitivity table showing IRR and MoM across ±1.0x exit multiple range
            \item Enables live iterative calculation in the interest expense calculation. Create a functional toggle that links to the interest expense line item
        \end{itemize}

        \item \textbf{Financing Terms}
        \begin{itemize}
            \item Revolver: \$2,000M capacity, undrawn at close, SOFR + 300bps, 1\% commitment fee, no OID, no financing fee
            \item 1st Lien TL: ~\$12,000M, SOFR + 450bps, 99 OID (1\% discount)
            \item 2nd Lien / HY: ~\$6,000M, SOFR + 600bps, 100 OID (par)
            \item Assume 1st and 2nd lien debt incur 2.00\% fee in financing fees to each tranche and no financing fees on revolver
            \item Build a fully functioning revolver that sweeps from minimum cash
            \item Debt repayment priority: FCF after minimum cash sweeps to 1st Lien TL until retired, then to 2nd Lien/HY
            \item Revolver draws only when cash falls below minimum cash and repays before any term debt prepayment
        \end{itemize}

    \item \textbf{Operating Assumptions}
    \begin{itemize}
        \item Forecast period: FY2027-FY2031 (5 years). FY2026 is the close year
        \item For forecast assumptions, use your own judgment anchored to historical trends, Electronic Arts’ disclosed guidance, and the business context in this prompt. Document your rationale for every forecast assumption
        \item 25\% effective tax rate. Cash taxes reflect deductibility of interest expense
        \item Model full interest tax shield
        \item Show required line items in the income statement, EBITDA build, and free cash flow bridge
        \item Develop your own financial forecast based off of your judgement and any management guidance
        \item IMPORTANT: Explicitly document rationale for each projected line item 
        \item Show FY2024A, FY2025A, FY2026E as historical/bridge columns
        \item Historical actuals, estimates, and projections should be in side by side columns
    \end{itemize}
    \end{enumerate}

    \item \textbf{Investment Memo (In PPT, bullet points preferred)}\\
    This memo is your formal recommendation to the Private Equity Mega Fund’s Investment Committee. It synthesizes the business case, financial analysis, and risk assessment into a single document that Investment Committee members will review in advance and use as the basis for discussion during the committee meeting. This document will serve as the basis for investment committee discussions and ultimately determines if this deal will move forward. The memo must be thorough enough to preemptively address Investment Committee questions, concise enough to be absorbed in 15-20 minutes, and structured to lead the committee to a clear go/no-go decision on whether to proceed with a binding offer. You may use the web to perform research on Electronic Arts and the industry. The investment memo should cover executive summary, business assessment, market analysis, investment thesis, deal terms and rationale, risks and mitigants, and key diligence items

\end{enumerate}
\end{PromptBoxBreak}
\captionof{figure}{Complete task definition for the LBO modeling task on Electronic Arts.}
\label{fig:lbo_rubric_full}

{\scriptsize\ttfamily
\setlength{\tabcolsep}{2pt}
\arrayrulecolor{rust}
\begin{longtable}{|p{0.12\textwidth}|p{0.14\textwidth}|p{0.06\textwidth}|p{0.68\textwidth}|}
\toprule
\rowcolor{gray!30} \textbf{Category} & \textbf{Sub-Category} & \textbf{Score} & \textbf{Criteria Description} \\
\toprule
\endfirsthead
\multicolumn{4}{c}{{LBO Task Rubric -- continued}} \\
\toprule
\rowcolor{gray!30} \textbf{Category} & \textbf{Sub-Category} & \textbf{Score} & \textbf{Criteria Description} \\
\midrule
\endhead
\midrule
\multicolumn{4}{r}{{Continued on next page}} \\
\endfoot
\bottomrule
\endlastfoot

 \multirow{2}{\linewidth}{\textbf{Pre-Scoring Gating Condition}} & \multirow{2}{\linewidth}{Correct Filing and Correct Period} & PASS & PASS = Proceed to Scoring: Correct filing types used; fiscal year-end correctly identified; No mixed periods or bases \\ \cline{3-4} \addlinespace[2pt]
 &  & FAIL & FAIL = Automatic 0: Wrong filing used; Wrong fiscal period or year; Mixed periods or bases; Fiscal year-end misidentified (Mixing FY with CY) \\ \cmidrule{1-4} \addlinespace[2pt]

\multirow{10}{\linewidth}{\textbf{Data Retrieval and Fidelity}} & \multirow{5}{\linewidth}{Source Compliance} & 50 & All cited or referenced material comes exclusively from permitted sources (SEC filings and permitted IR website); each reference explicitly identifies the source type and fiscal period \\ \cline{3-4} \addlinespace[2pt]
 & & 20 & All material appears to come from permitted sources; one reference is imprecisely labeled but the intended source is clear \\ \cline{3-4} \addlinespace[2pt]
 & & 10 & Multiple references are ambiguously labeled, but there is no evidence of non-permitted source usage or hallucinated data \\ \cline{3-4} \addlinespace[2pt]
 & & 5 & One instance of language that implicitly suggests use of non-permitted sources without explicit citation \\ \cline{3-4} \addlinespace[2pt]
 & & 0 & Explicit use of non-permitted sources, fabricated numbers, or hallucinated financial data with no traceable origin \\ \cmidruleFull \addlinespace[2pt]
 
 & \multirow{5}{\linewidth}{Numerical Accuracy Check} & 50 & All historical values exactly match SEC filings, no rounding, no sign errors, no normalization or transformation, and values are clearly sourced to the correct filing, fiscal year, and financial statement \\ \cline{3-4} \addlinespace[2pt]
 & & 40 & One or more immaterial numerical discrepancies such as rounding or sign convention issues limited to non-core line items but all values are clearly derived from filings \\ \cline{3-4} \addlinespace[2pt]
 & & 20 & One or more material numerical errors / missing value affecting non-core line items, or multiple small discrepancies that reduce confidence but do not affect core metrics \\ \cline{3-4} \addlinespace[2pt]
 & & 10 & One or more material numerical errors / missing values affecting core line items such as revenue, COGS, Net Income, or widespread numerical inaccuracies despite filing-derived values \\ \cline{3-4} \addlinespace[2pt]
 & & 0 & Systematic numerical issues such as consistent rounding, normalization, aggregation, or partial estimation, requiring substantial re-extraction, large portion of numerical inputs do not match historicals \\ \cmidrule{1-4} \addlinespace[2pt]

\multirow{20}{\linewidth}{\textbf{Transaction Structure and Valuation}} & \multirow{4}{\linewidth}{Entry Valuation Build} & 25 & Offer price x FDSO = Equity Value computed correctly; FDSO calculation shows full work; TSM formula is correct if options are in the money(if Options are applicable to the deal); all components visible and traceable \\ \cline{3-4} \addlinespace[2pt]
 &  & 15 & Equity Value is correct; FDSO calculation present but TSM has a minor error (wrong exercise price or share price used); overall result is approximately right \\ \cline{3-4} \addlinespace[2pt]
 &  & 10 & Equity Value is approximately right but FDSO is hardcoded without showing the TSM work; reviewer cannot verify the share count independently \\ \cline{3-4} \addlinespace[2pt]
 &  & 0 & Equity Value is materially wrong; FDSO is incorrect or absent; entry valuation cannot be relied upon \\ \cmidruleFull \addlinespace[2pt]
 
 & \multirow{4}{\linewidth}{Equity-to- Enterprise Value Bridge} & 30 & Bridge from Equity Value to Enterprise Value is correct with all components (less cash, plus total debt, plus any other adjustments) pulled from the closing balance sheet; all items sourced from the correct period \\ \cline{3-4} \addlinespace[2pt]
 &  & 20 & Bridge is correct in structure but missing a minor component (eg. minority interest omitted when immaterial) or uses a slightly different period for one item \\
 &  & 5 & Bridge is directionally right but uses the wrong period for major components (eg. FY2025 balance sheet instead of FY2026E) or is missing a major component like total debt \\ \cline{3-4} \addlinespace[2pt]
 &  & 0 & Bridge is materially wrong or absent; Enterprise Value cannot be reconciled from the Equity Value \\ \cmidruleFull \addlinespace[2pt]
 
 & \multirow{4}{\linewidth}{Implied Multiples} & 15 & LTM EBITDA multiple and Net Debt/EBITDA at entry are both shown, formula-driven from the bridge and EBITDA values, not hardcoded \\ \cline{3-4} \addlinespace[2pt]
 & & 10 & Multiples are shown but at least one is hardcoded rather than formula-linked to the bridge \\ \cline{3-4} \addlinespace[2pt]
 & & 5 & More than one multiple is hardcoded, or multiples reference the wrong EBITDA period \\ \cline{3-4} \addlinespace[2pt]
 & & 0 & Implied multiples are absent or materially incorrect \\ \cmidruleFull \addlinespace[2pt]
 
 & \multirow{4}{\linewidth}{Deal Assumptions Presentation} & 25 & All deal assumptions (offer price, FDSO, advisory fees, minimum cash, tax rate, SOFR, exit multiple) presented in blue font as hardcoded inputs in a clearly labeled, centralized assumptions section; no embedded constants in any formula \\ \cline{3-4} \addlinespace[2pt]
 & & 15 & Most assumptions are centralized in a labeled section with blue font, but some constants are embedded in formulas or a few assumptions are scattered outside the main block \\ \cline{3-4} \addlinespace[2pt]
 & & 5 & Assumptions are scattered across the model; difficult to identify where inputs live; limited use of blue font convention \\ \cline{3-4} \addlinespace[2pt]
 & & 0 & Assumptions are not identifiable as a distinct section; inputs are embedded throughout with no visual distinction or labeling \\ \cmidruleFull \addlinespace[2pt]
 
 & \multirow{4}{\linewidth}{Debt Sizing and Terms} & 25 & All debt tranches correctly sized from the correct LTM EBITDA anchor; rates (SOFR + spread), OID, commitment fees, and financing fees all match the prompt specifications exactly \\ \cline{3-4} \addlinespace[2pt]
 & & 15 & Debt sizing is correct but one financing term is wrong (eg. wrong spread on a tranche, incorrect OID, or wrong financing fee percentage) \\ \cline{3-4} \addlinespace[2pt]
 & & 5 & Multiple financing terms are incorrect or inconsistent with the prompt \\ \cline{3-4} \addlinespace[2pt]
 & & 0 & Debt structure is materially wrong (wrong leverage multiples, wrong tranche sizes, or missing a tranche entirely) \\ \cmidrule{1-4} \addlinespace[2pt]

\multirow{8}{\linewidth}{\textbf{Sources and Uses}} & \multirow{4}{\linewidth}{Sources Completeness} & 25 & Sources include: 1L TL, 2L/HY (if applicable), Sponsor Equity; Revolver correctly excluded (undrawn at transaction close) \\ \cline{3-4} \addlinespace[2pt]
 & & 10 & All major sources present but one source component is mislabeled \\ \cline{3-4} \addlinespace[2pt]
 & & 5 & Sources are partially complete; one major component is missing or Revolver is incorrectly included as a source \\ \cline{3-4} \addlinespace[2pt]
 & & 0 & Sources section is materially incomplete or structurally wrong \\ \cmidruleFull \addlinespace[2pt]

 & \multirow{4}{\linewidth}{Uses Completeness} & 25 & Uses include: Purchase Equity Value, Refinancing of Existing Net Debt, Cash to Balance Sheet, Transaction Advisory Fees, Financing Fees; all fees calculated correctly from the right base \\ \cline{3-4} \addlinespace[2pt]
 &  & 10 & All major uses present but one fee calculation has a minor error (eg. financing fees computed on net rather than gross debt) \\ \cline{3-4} \addlinespace[2pt]
 &  & 5 & Uses are partially complete; one major component (eg. advisory fees or financing fees) is missing entirely \\ \cline{3-4} \addlinespace[2pt]
 &  & 0 & Uses section is materially incomplete or structurally wrong \\ \cmidrule{1-4} \addlinespace[2pt]

\multirow{29}{\linewidth}{\textbf{Financial Forecast and Cash Flow}} & \multirow{3}{\linewidth}{Last Historical Year Bridge Schedule} & 30 & Bridge schedule is present below sensitivity tables with YTD actuals, projected remaining quarter(s), and full-year totals clearly separated; projection uses averaged QoQ sequential growth rates from the prior two fiscal years applied to the most recent actual quarter; main forecast section links directly to bridge totals with no recomputation or hardcoded values; results are not annualized \\ \cline{3-4} \addlinespace[2pt]
 & & 10 & Bridge exists but the projected quarter is derived using a different method than specified (such as annualization, flat assumption, or share of year ratio); or main forecast recomputes figures instead of linking to the bridge \\ \cline{3-4} \addlinespace[2pt]
 & & 0 & No bridge schedule; stub year figures are hardcoded, annualized, or absent entirely \\ \cmidruleFull \addlinespace[2pt]
 
 & \multirow{3}{\linewidth}{Guidance Integration} & 30 & Most recent 8k demonstrably read; specific management guidance extracted (eg. revenue guidance, live services growth trajectory, margin commentary) and directly applied to forecast assumptions with explicit linkage \\ \cline{3-4} \addlinespace[2pt]
 & & 10 & Generic reference to management guidance; guidance loosely referenced but not mapped to specific forecast assumptions \\ \cline{3-4} \addlinespace[2pt]
 & & 0 & No evidence of 8k review; forecast assumptions show no connection to disclosed management guidance \\ \cmidruleFull \addlinespace[2pt]
 
 & \multirow{4}{\linewidth}{Revenue Forecast Quality} & 15 & Revenue is built with segment-level detail or clearly identifiable growth drivers; growth rates are anchored to historical trends and management guidance with documented rationale \\ \cline{3-4} \addlinespace[2pt]
 & & 10 & Revenue is forecast at a consolidated level but growth rates are well-reasoned with explicit rationale tied to business drivers \\ \cline{3-4} \addlinespace[2pt]
 & & 5 & Revenue growth is flat or arbitrary with no documented rationale; growth rates do not reflect business context or historical patterns \\ \cline{3-4} \addlinespace[2pt]
 & & 0 & Revenue forecast is nonsensical (extreme unexplained jumps or declines) or entirely absent \\ \cmidruleFull \addlinespace[2pt]
 
 & \multirow{4}{\linewidth}{Margin and Expense Forecast} & 15 & Gross margin, SG\&A, R\&D are each driven by historical trends with PE-lens adjustments (cost takeout, margin expansion thesis); margins as \% of revenue shown as blue-font operating assumptions; rationale documented for deviations from historical levels \\ \cline{3-4} \addlinespace[2pt]
 & & 10 & Major expense lines are present and generally reasonable, but some rely on simplified assumptions (eg. flat margin with no thesis for improvement); limited PE-specific adjustments \\ \cline{3-4} \addlinespace[2pt]
 & & 5 & Expense forecast is generic (eg. all items as fixed \% of revenue with no historical grounding); no margin expansion or cost thesis articulated \\ \cline{3-4} \addlinespace[2pt]
 & & 0 & Expense forecast is absent, nonsensical, or internally inconsistent \\ \cmidruleFull \addlinespace[2pt]

 & \multirow{4}{\linewidth}{EBITDA Build} & 25 & Adjusted EBITDA is built up correctly (EBIT + D\&A + SBC + one time adjustments); all add-backs are visible, labeled, and individually traceable; EBITDA build reconciles cleanly to the income statement components above it \\ \cline{3-4} \addlinespace[2pt]
 & & 15 & EBITDA build is present and mostly correct but missing one add-back (eg. SBC omitted) or has a minor reconciliation error that does not materially distort the result \\ \cline{3-4} \addlinespace[2pt]
 & & 5 & EBITDA is computed as a margin assumption applied to revenue (no build-up from components); cannot trace EBITDA to the underlying P\&L \\ \cline{3-4} \addlinespace[2pt]
 & & 0 & EBITDA build is wrong, absent, or the number used downstream does not reconcile to any visible calculation \\ \cmidruleFull \addlinespace[2pt]

 & \multirow{4}{\linewidth}{EBITDA to Free Cash Flow} & 30 & FCF build follows the correct sequence: EBITDA, less D\&A = EBIT, less interest expense (linked from debt schedule), less taxes, plus D\&A add-back, plus SBC, plus/minus NWC changes, less CapEx = FCF (before debt repayment); visible FCF check cell confirms the calculation; all components are formula-driven \\ \cline{3-4} \addlinespace[2pt]
 & & 20 & FCF flow is structurally correct with a minor error in one component (eg. NWC sign convention reversed); overall FCF is approximately right and the logic is traceable \\ \cline{3-4} \addlinespace[2pt]
 & & 10 & FCF is calculated but some components are hardcoded or missing (eg. NWC changes omitted, CapEx assumed rather than linked); no FCF check cell \\ \cline{3-4} \addlinespace[2pt]
 & & 0 & FCF logic is broken; cash flow cannot be traced from EBITDA through to the debt schedule; or FCF section is absent \\ \cmidruleFull \addlinespace[2pt]

 & \multirow{4}{\linewidth}{Interest Expense Linkage} & 30 & Interest expense in the forecast is pulled directly from the debt schedule (not hardcoded); reflects circularity (interest depends on debt balance, which depends on FCF, which depends on interest); circularity toggle at top of sheet is functional (0 = interest zeroed, 1 = live iterative calc) \\ \cline{3-4} \addlinespace[2pt]
 & & 15 & Interest expense is linked from the debt schedule but circularity is not handled (no toggle, or toggle exists but does not function); model may produce circular reference errors \\ \cline{3-4} \addlinespace[2pt]
 & & 5 & Interest expense is hardcoded in the forecast rather than linked from the debt schedule; changes to debt balances do not flow through to interest \\ \cline{3-4} \addlinespace[2pt]
 & & 0 & Interest expense is absent, wrong, or the linkage between forecast and debt schedule is broken \\ \cmidruleFull \addlinespace[2pt]

 & \multirow{4}{\linewidth}{Operating Assumptions Section} & 25 & All forecast drivers (revenue growth \%, gross margin \%, SG\&A \% revenue, R\&D \% revenue, D\&A \% revenue, CapEx \% revenue, NWC \% revenue, SBC, tax rate) shown in a clearly labeled assumptions block in blue font; changing one assumption flows through the entire model dynamically \\ \cline{3-4} \addlinespace[2pt]
 & & 15 & Most forecast drivers are centralized in an assumptions block with blue font, but a few are embedded in formulas or not all drivers are shown explicitly \\ \cline{3-4} \addlinespace[2pt]
 & & 5 & Assumptions are scattered; limited blue font usage; changing an assumption does not reliably propagate through the model \\ \cline{3-4} \addlinespace[2pt]
 & & 0 & No identifiable operating assumptions section; forecast drivers are embedded or hardcoded throughout \\ \cmidrule{1-4} \addlinespace[2pt]

\multirow{19}{\linewidth}{\textbf{Debt Schedule and Mechanics}} & \multirow{4}{\linewidth}{Debt Schedule Structure} & 20 & Debt schedule shows beginning balance, mandatory amortization, optional prepayment, ending balance, interest rate, and interest expense for each tranche (Revolver, 1L TL, 2L/HY) in each forecast year (FY2027-FY2031); structure is clean and auditable \\ \cline{3-4} \addlinespace[2pt]
 &  & 10 & Debt schedule is present for all tranches but missing one column (eg. no separate mandatory amort vs. optional prepayment breakdown) or one tranche is partially incomplete \\ \cline{3-4} \addlinespace[2pt]
 &  & 5 & Debt schedule exists but is structurally incomplete (eg. only shows ending balances without the roll-forward, or missing a tranche entirely) \\ \cline{3-4} \addlinespace[2pt]
 &  & 0 & Debt schedule is absent or non-functional; no tranche-level detail \\ \cmidruleFull \addlinespace[2pt]

 & \multirow{3}{\linewidth}{Mandatory Amortization} & 20 & 1L TL mandatory amort is calculated correctly (x\% per annum of original principal); amort is applied before the cash sweep in the waterfall; 2L/HY correctly shows no mandatory amortization. Correct amortization logic if prompt doesn't require amorization \\ \cline{3-4} \addlinespace[2pt]
 & & 10 & Mandatory amort exists but the rate is wrong, or it is applied in the wrong order relative to the cash sweep; or 2L incorrectly includes amort \\ \cline{3-4} \addlinespace[2pt]
 & & 0 & No mandatory amortization modeled on any tranche when prompt specified calculations as required; or amort logic is fundamentally broken \\ \cmidruleFull \addlinespace[2pt]
 
 & \multirow{4}{\linewidth}{Cash Sweep Mechanics} & 30 & Full cash sweep waterfall working correctly: (1) starts with FCF before debt repayment, (2) subtracts mandatory amort (if prompt mentions), (3) checks against minimum cash, (4) sweeps excess to 1L TL until retired, (5) then sweeps to 2L/HY; ending cash equals minimum cash (or higher only if all debt is fully repaid) \\  \cline{3-4} \addlinespace[2pt]
 & & 20 & Cash sweep works and generally follows the waterfall but priority order is slightly off (eg. sweeps to 2L before 1L is fully retired) or minimum cash check has a minor error \\ \cline{3-4} \addlinespace[2pt]
 & & 10 & Cash sweep mechanism exists but does not handle tranche priority (eg. pro-rata paydown rather than sequential) or does not correctly enforce minimum cash \\ \cline{3-4} \addlinespace[2pt]
 & & 0 & No cash sweep modeled; debt balances are static or manually adjusted; or sweep logic is fundamentally broken \\ \cmidruleFull \addlinespace[2pt]
 
 & \multirow{4}{\linewidth}{Revolver Mechanics} & 25 & Revolver draws dynamically when cash falls below minimum cash; repays before any term debt prepayment; correctly caps at maximum capacity; commitment fee calculated on undrawn balance; interest calculated only on drawn balance; revolver is fully functional across all scenarios \\ \cline{3-4} \addlinespace[2pt]
 & & 15 & Revolver draws and repays dynamically but fee calculations have an error (commitment fee on total capacity rather than undrawn, or interest on capacity rather than drawn amount) \\ \cline{3-4} \addlinespace[2pt]
 & & 5 & Revolver exists in the model but does not function dynamically (eg. always shows zero balance regardless of cash position, or is hardcoded) \\ \cline{3-4} \addlinespace[2pt]
 & & 0 & Revolver is absent or purely decorative (labeled but no formulas); model has no mechanism for short-term liquidity management \\ \cmidruleFull \addlinespace[2pt]
 
 &  \multirow{4}{\linewidth}{Interest Calculation Accuracy} & 30 & Interest on each tranche calculated as average balance x (SOFR + spread); total interest expense includes 1L interest + 2L interest (if applicable) + revolver interest (if drawn) + commitment fee on undrawn revolver + amortization of financing fees (total fees / 5 years); cash interest income accounted for \\ \cline{3-4} \addlinespace[2pt]
 &  & 10 & Interest calculations are mostly correct but one component is wrong (eg. uses ending balance instead of average, or financing fee amortization is missing) \\ \cline{3-4} \addlinespace[2pt]
 & & 5 & Interest is calculated but multiple components are wrong or simplified (eg. flat interest assumption rather than balance-linked, missing commitment fee and financing fee amort) \\ \cline{3-4} \addlinespace[2pt]
 & & 0 & Interest calculations are fundamentally wrong or hardcoded with no connection to debt balances \\ \cmidruleFull \addlinespace[2pt]

 \multirow{15}{\linewidth}{\textbf{Returns and Sensitivity}} &  \multirow{4}{\linewidth}{Returns Calculation Structure} & 25 & Model shows LTM Adjusted EBITDA x Exit Multiple = TEV, less Net Debt = Equity Value for each forecast year; initial equity investment shown as negative cash flow at entry; IRR calculated using XIRR or IRR function with correct dates/periods; MoM calculated as exit equity / entry equity; all components formula-driven \\ \cline{3-4} \addlinespace[2pt]
 & & 15 & Returns structure is correct but one formula has an error (eg. IRR dates are off by one period, or net debt in the exit calc does not match the debt schedule ending balance) \\ \cline{3-4} \addlinespace[2pt]
 & & 5 & Returns are shown but methodology is unclear (eg. IRR calculated manually rather than with a function, or exit equity value does not trace to TEV minus net debt) \\
 & & 0 & Returns calculation is absent or fundamentally wrong (eg. no exit value computed, IRR/MoM not calculated) \\ \cmidruleFull \addlinespace[2pt]
 
 &  \multirow{3}{\linewidth}{IRR and MoM Display and Linkage} & 20 & 5-year IRR and MoM are displayed in the Transaction Drivers section at the top of the sheet; both are formula-linked to the Returns Calculation section below; values update dynamically when any assumption changes \\ \cline{3-4} \addlinespace[2pt]
 & & 10 & IRR and MoM are displayed at the top but one is hardcoded or the link to the returns section is broken; or they are only shown in the returns section, not at the top \\ \cline{3-4} \addlinespace[2pt]
 & & 0 & IRR and MoM are not displayed at the top of the sheet; or both are hardcoded; or neither metric is present \\ \cmidruleFull \addlinespace[2pt]
 
 &  \multirow{4}{\linewidth}{Sensitivity Tables} & 20 & Two sensitivity tables present: IRR and MoM, both varying entry premium vs. exit multiple; formula-driven (DATA TABLE function or equivalent); base case sits in the correct cell of the matrix and is visually highlighted \\ \cline{3-4} \addlinespace[2pt]
 & & 10 & Sensitivity tables are present but partially hardcoded, or only one metric (IRR or MoM) is shown, or base case is not correctly positioned in the matrix \\ \cline{3-4} \addlinespace[2pt]
 & & 5 & Tables are present but wrong axes are used, or tables are broken (formulas do not update when inputs change) \\ \cline{3-4} \addlinespace[2pt]
 & & 0 & Sensitivity tables are absent \\ \cmidruleFull \addlinespace[2pt]
 
 &  \multirow{4}{\linewidth}{Sensitivity Range Reasonableness} & 15 & Ranges are commercially sensible \\ \cline{3-4} \addlinespace[2pt]
 & & 10 & Ranges are present and mostly reasonable but slightly narrow or wide; directional relationships are correct \\ \cline{3-4} \addlinespace[2pt]
 & & 5 & Ranges are present but include unreasonable values (eg. 0\% premium or 25x exit multiple) or directional relationships are inverted in parts of the table \\ \cline{3-4} \addlinespace[2pt]
 & & 0 & Ranges are absent, nonsensical, or sensitivity outputs are clearly wrong (eg. IRR increases as premium increases) \\ \cmidrule{1-4} \addlinespace[2pt]

 \multirow{14}{\linewidth}{\textbf{Model Engineering and Reviewability}} & \multirow{4}{\linewidth}{Formatting Compliance} & 15 & Blue font = hardcoded input, Black font = formula, Green font = cross-section link consistently applied throughout; Navy fill with white bold text on all section headers; formatting is clean and consistent \\ \cline{3-4} \addlinespace[2pt]
 & & 10 & Color convention is mostly followed but inconsistent in places (eg. some inputs in black, some section headers not navy); overall intent is clear \\ \cline{3-4} \addlinespace[2pt]
 & & 5 & Limited adherence to color coding; some section headers formatted but font colors are largely inconsistent \\ \cline{3-4} \addlinespace[2pt]
 & & 0 & No discernible formatting convention; inputs, formulas, and links are visually indistinguishable \\ \cmidruleFull \addlinespace[2pt]
 
 & \multirow{3}{\linewidth}{Circularity Toggle} & 20 & Circularity toggle cell is present at the top of the sheet; setting to 0 zeros out interest expense and breaks the circular reference; setting to 1 enables live iterative calculation; toggle is functional and tested \\ \cline{3-4} \addlinespace[2pt]
 & & 10 & IF logic is present across the forecast years but is not linked to the toggle cell and/or toggle cell is empty (NOT 0 or 1) \\ \cline{3-4} \addlinespace[2pt]
 & & 0 & No circularity toggle; model either has uncontrolled circular references or avoids circularity by hardcoding interest \\ \cmidruleFull \addlinespace[2pt]
 
 & \multirow{2}{\linewidth}{S\&U Balance Check} & 20 & Sources = Uses exactly; visible balance check cell near the top of the sheet confirms this; Sponsor Equity is calculated as the residual plug \\ \cline{3-4} \addlinespace[2pt]
 & & 0 & Sources do not equal Uses; balance check is absent or shows an error; or Sponsor Equity is not the plug \\ \cmidruleFull \addlinespace[2pt]
 
 & \multirow{5}{\linewidth}{Calculation Transparency and Auditability} & 30 & All calculated values clearly expose components and intermediate steps; calculation logic is readable without formula inspection; model is logically organized top-to-bottom (Transaction Drivers > Assumptions > S\&U > Forecast > Debt Schedule > Returns > Sensitivity); no hidden hardcodes, forced plugs, or manual overrides; any number can be traced to its source quickly \\ \cline{3-4} \addlinespace[2pt]
 & & 20 & Model is largely auditable with clear flow and traceability; minor issues such as limited hardcoding in non-critical areas, or one section out of standard order \\ \cline{3-4} \addlinespace[2pt]
 & & 10 & Auditability is mixed but structural integrity exists; frequent hardcodes, scattered assumptions, or partially opaque formulas that materially slow validation but do not prevent it \\ \cline{3-4} \addlinespace[2pt]
 & & 5 & Model is difficult to audit due to unclear structure, hidden or inconsistent logic, weak reconciliation checks, or widespread hardcoding \\ \cline{3-4} \addlinespace[2pt]
 & & 0 & Model is not meaningfully auditable; hidden overrides, forced plugs, broken logic, or structural opacity that prevents reliable validation \\ \cmidrule{1-4} \addlinespace[2pt]

\multirow{23}{\linewidth}{\textbf{Investment Memo}} & \multirow{3}{\linewidth}{Executive Summary and Investment Thesis} & 30 & Transaction snapshot (target, offer price, equity value, enterprise value, entry multiple) is complete; 3-5 distinct, defensible reasons to invest are articulated; headline returns (IRR, MoM) are included; clear go/no-go recommendation from the deal team \\ \cline{3-4} \addlinespace[2pt]
 &  & 15 & Executive summary is present with most key elements but missing one component (eg. no explicit recommendation, or entry multiple not stated); thesis points are reasonable but not all are fully developed \\ \cline{3-4} \addlinespace[2pt]
 & & 0 & Executive summary is generic or boilerplate or absent; investment thesis is vague ("XYZ is a good company") without specific, defensible arguments \\ \cmidruleFull \addlinespace[2pt]
 
 & \multirow{4}{\linewidth}{Company and Business Overview} & 40 & Thorough analysis of business model (revenue streams, platform mix, geographic split), key franchise portfolio with durability assessment, competitive positioning vs. named peers, and management team assessment; supported by data from filings and industry research \\ \cline{3-4} \addlinespace[2pt]
 & & 20 & Business overview is solid but surface-level; covers the major points without deep analysis of franchise durability or competitive dynamics; management assessment is thin or absent \\ \cline{3-4} \addlinespace[2pt]
 & & 5 & The business overview could apply to any company in the same industry; no company specific analysis and no data from filings \\ \cline{3-4} \addlinespace[2pt]
 & & 0 & Business overview is absent or contains materially incorrect information about the company \\ \cmidruleFull \addlinespace[2pt]

 & \multirow{3}{\linewidth}{Industry and Market Context} & 20 & Covers global market size and growth, consolidation trends with specific comparable transactions cited, regulatory landscape, and secular risks (platform shifts, AI disruption) \\ \cline{3-4} \addlinespace[2pt]
 & & 10 & Industry overview is present but incomplete; mentions market size without specific data, or omits regulatory or secular risk discussion \\ \cline{3-4} \addlinespace[2pt]
 & & 0 & Industry context is absent or generic boilerplate with no industry specific analysis \\ \cmidruleFull \addlinespace[2pt]
 
 & \multirow{3}{\linewidth}{Deal Terms, Valuation, and Capital Structure} & 30 & Evaluates entry valuation in context (implied multiples vs. public comps and precedent transactions); analyzes offer premium relative to historical premiums in comparable take-privates; discusses capital structure in context of recent mega-LBOs; sources and uses summary included; articulates value creation thesis (deleveraging, margin expansion, organic growth) with return attribution \\ \cline{3-4} \addlinespace[2pt]
 & & 10 & Deal terms are presented clearly but benchmarking is limited (eg. states the entry multiple without comparing to comps, or mentions leverage without context from comparable deals) \\ \cline{3-4} \addlinespace[2pt]
 & & 0 & Deal terms are restated from the prompt without analytical context; no comp benchmarking or value creation thesis \\ \cmidruleFull \addlinespace[2pt]
 
 & \multirow{3}{\linewidth}{Returns Analysis and Sensitivity Discussion} & 30 & Presents base, upside, and downside return scenarios with clearly stated assumptions; references sensitivity analysis from the model; discusses key assumption stress tests and how returns degrade under adverse conditions \\ \cline{3-4} \addlinespace[2pt]
 & & 5 & Returns are mentioned but only base case is discussed; no scenario analysis or sensitivity discussion \\ \cline{3-4} \addlinespace[2pt]
 & & 0 & Returns discussion is absent or only restates model outputs without analysis \\ \cmidruleFull \addlinespace[2pt]
 
 & \multirow{3}{\linewidth}{Diligence Priorities and Risk Assessment} & 20 & Identifies specific risks across product execution (franchise fatigue, pipeline), regulatory exposure, leverage sustainability (downside), and post-privatization cost structure (SBC cash replacement, headcount); each risk is paired with a specific data request or diligence workstream; risks are company specific, not generic \\ \cline{3-4} \addlinespace[2pt]
 & & 10 & Key risks are identified and mostly company specific, but data requests are vague \\ \cline{3-4} \addlinespace[2pt]
 & & 5 & Risks are generic ("market risk," "execution risk") without company specific detail; some major risk categories are missing (eg. no discussion of regulatory risk or leverage sustainability) and/or only includes generic risk factors; no data requests specified \\ \cline{3-4} \addlinespace[2pt]
 & & 0 & Risk assessment is absent or boilerplate with no connection to the company or the deal structure \\ \cmidruleFull \addlinespace[2pt]
 
 & \multirow{3}{\linewidth}{Memo Quality, Structure, and Presentation} & 10 & Memo is well organized across all required pages/sections; bullet-point format as specified; concise but comprehensive; IC-ready quality (a senior partner could read it in 15-20 minutes and make a decision); no factual errors or internal inconsistencies \\ \cline{3-4} \addlinespace[2pt]
 & & 5 & Memo structure is adequate but some sections are uneven in depth; occasional redundancy; a few minor factual inconsistencies \\ \cline{3-4} \addlinespace[2pt]
 & & 0 & Memo is disorganized, incomplete, or not in the specified format; would not be presentable to an IC \\ 
\end{longtable}
\caption{Detailed evaluation rubric for LBO modeling tasks} 
\label{fig:lbo_rubric_detailed}
}

\end{document}